\documentclass[a4paper,fleqn]{cas-sc}

\usepackage[authoryear,longnamesfirst]{natbib}

\usepackage{graphicx}
\usepackage{lscape}
\usepackage{caption}
\usepackage{subfig}
\usepackage{float}
\usepackage{adjustbox}
\usepackage{booktabs}
\usepackage{enumitem}

\usepackage{algorithm}
\usepackage{algpseudocode}

\usepackage{amsmath}
\usepackage{bbm}          
\usepackage{gensymb}

\usepackage{xcolor}
\usepackage{colortbl}

\def\tsc#1{\csdef{#1}{\textsc{\lowercase{#1}}\xspace}}
\tsc{WGM}
\tsc{QE}
\tsc{EP}
\tsc{PMS}
\tsc{BEC}
\tsc{DE}

\begin{document}

\let\WriteBookmarks\relax
\def\floatpagepagefraction{1}
\def\textpagefraction{.001}

\shorttitle{Enhancing Dual Network Based Semi-Supervised Medical Image Segmentation with Uncertainty-Guided Pseudo-Labeling}

\shortauthors{Lu et al.}

\title [mode=title]{Enhancing Dual Network Based Semi-Supervised Medical Image Segmentation with Uncertainty-Guided Pseudo-Labeling}

\fntext[fn1]{Yunyao Lu, Yihang Wu and Ahmad Chaddad are contributed equally to this work.}
\author[add1]{Yunyao~Lu}
\author[add1]{Yihang~Wu}
\author[add1,add2]{Ahmad Chaddad \corref{cor1}}[orcid=0000-0003-3402-9576]
\ead{ahmadchaddad@guet.edu.cn}
\author[add3, add4]{Tareef~Daqqaq}
\author[add5,add6]{Reem~Kateb}



\address[add1]{School of Artificial Intelligence, Guilin University of Electronic Technology, Guilin, China, 541004}
\address[add2]{The Laboratory for Imagery, Vision and Artificial Intelligence, École de Technologie Supérieure, Montreal, Canada, H3C 1K3}
\address[add3]{College of Medicine, Taibah University, Al Madinah, Saudi Arabia, 42353}
\address[add4]{Department of Radiology, Prince Mohammed Bin Abdulaziz Hospital, Ministry of National Guard Health Affairs, Al Madinah, Saudi Arabia,  42324}
\address[add5]{College of Computer Science and Engineering, Taibah University, Madinah, Saudi Arabia, 42353}
\address[add6]{College of Computer Science and Engineering, Jeddah University, Jeddah, Saudi Arabia, 23445}

\begin{abstract}
Despite the remarkable performance of supervised medical image segmentation models, relying on a large amount of labeled data is impractical in real-world situations. Semi-supervised learning approaches aim to alleviate this challenge using unlabeled data through pseudo-label generation. Yet, existing semi-supervised segmentation methods still suffer from noisy pseudo-labels and insufficient supervision within the feature space. To solve these challenges, this paper proposes a novel semi-supervised 3D medical image segmentation framework based on a dual-network architecture. Specifically, we investigate a Cross Consistency Enhancement module using both cross pseudo and entropy-filtered supervision to reduce the noisy pseudo-labels, while we design a dynamic weighting strategy to adjust the contributions of pseudo-labels using an uncertainty-aware mechanism (i.e., Kullback–Leibler divergence). In addition, we use a self-supervised contrastive learning mechanism to align uncertain voxel features with reliable class prototypes by effectively differentiating between trustworthy and uncertain predictions, thus reducing prediction uncertainty. Extensive experiments are conducted on three 3D segmentation datasets, Left Atrial, NIH Pancreas and
BraTS-2019. The proposed approach consistently exhibits superior performance across various settings (e.g., 89.95\% Dice score on left Atrial with 10\% labeled data) compared to the state-of-the-art methods. Furthermore, the usefulness of the proposed modules is further
validated via ablation experiments. The code repository is available at \url{https://github.com/AIPMLab/Semi-supervised-Segmentation}.

\end{abstract}


\maketitle

\section{Introduction}

Medical image segmentation is a fundamental step in computer-aided diagnosis, treatment planning, and surgical intervention \cite{yu2025edge}. It involves identifying and labeling pixels or voxels that correspond to anatomical or pathological structures within biomedical images \cite{loussaief2025adaptive,illarionova2025hierarchical}. By accurately delineating and quantifying these regions, segmentation provides essential diagnostic insights that support precision medicine \cite{caie2025precision} and improve the reliability of decision-making procedures \cite{zou2025segment}.
With the continuous advancement of deep learning (DL) technologies, numerous high-performance methods have been developed for medical image segmentation. Among them, the U-Net architecture \cite{ronneberger2015u} and its variants \cite{cciccek20163d,revathi2025brain} have become dominant approaches in medical image segmentation, thanks to their encoder–decoder structure and strong ability to capture fine-grained anatomical details. Such advanced DL frameworks have significantly improved segmentation accuracy, setting new benchmarks in the field. However, DL models are prone to overfitting when training data is scarce, making the lack of high-quality annotated data a major challenge in medical image segmentation \cite{sumon2025multiclass}. Most state-of-the-art (SOTA) segmentation models depend heavily on large-scale, pixel-level annotations, which are expensive and time-consuming to obtain due to the need for expert clinical knowledge \cite{wang2025learning}. To address this limitation, it is crucial to develop robust models that can learn effectively from limited labeled data supplemented by abundant unlabeled data \cite{peng2021medical}. As a result, semi-supervised learning (SSL) techniques \cite{chen2025multi}—designed to leverage both labeled and unlabeled data—have attracted increasing attention in medical image segmentation tasks.

\begin{figure}
    \centering
    \includegraphics[width=0.4\linewidth]{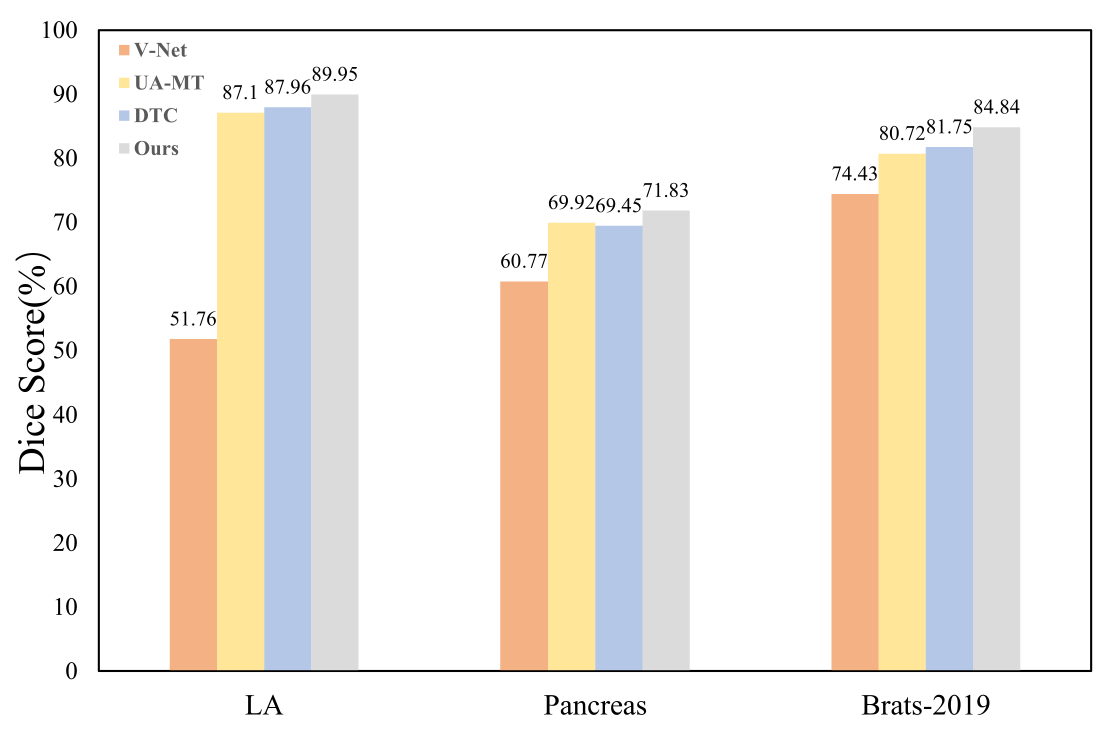} \includegraphics[width=0.49\linewidth]{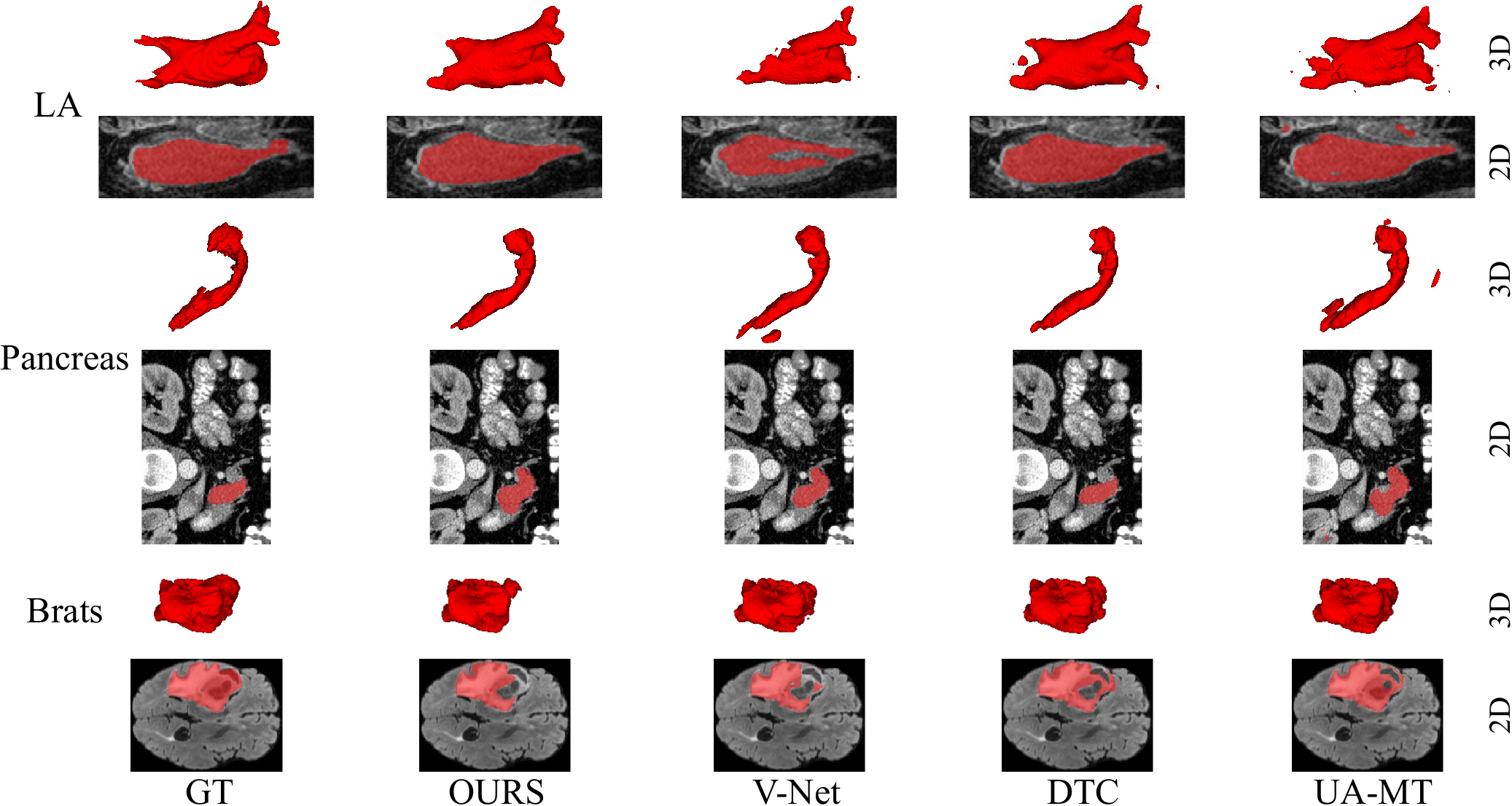}
    \caption{Quantitative (Dice score) and visualization comparisons between our method and recent semi-supervised approaches on the LA, Pancreas, and BraTS-2019 datasets using 10\% labeled data.}
    \label{fig:Visual}
\end{figure}

A common strategy in SSL is pseudo-supervision, where model-generated pseudo-labels are used to train on unlabeled data in the absence of ground truth annotations \cite{su2024mutual}. While pseudo-labeling has led to notable progress, semi-supervised medical image segmentation still faces two key challenges: (1) noisy pseudo-labels, and (2) insufficient supervision in the feature space. The generation of pseudo-labels depends on the model's current predictions, which are often unreliable, especially in the early stages of training due to limited labeled data. As a result, pseudo-labels can be noisy, and segmentation models, which are highly sensitive to label noise, suffer reduced performance under such conditions \cite{fooladgar2024manifold}. {To address label noise, recent studies employed cross pseudo supervision (CPS), using two networks to provide pseudo-labels for each other to refine predictions collaboratively \cite{chen2021semi, wang2023mcf, su2024mutual}. Furthermore, entropy-based filtering is explored to detect low-confidence predictions and retain only reliable supervision signals \cite{xu2023ambiguity, su2024mutual}. Prototype-guided learning is introduced to enforce intra-class compactness and inter-class separation in the embedding space \cite{xu2022all, karimijafarbigloo2024leveraging}. However, these techniques are often applied separately or simply integrated, limiting their full potential.} In parallel, many existing SSL methods focus primarily on supervising the label space (i.e., output predictions) while overlooking explicit guidance in the feature space. This lack of feature-level supervision limits the model ability to learn discriminative representations, leading to insufficient class separability \cite{ouali2020semi, chen2021semi, zhang2021flexmatch}. Although recent approaches attempt to reduce label noise by filtering out low-confidence predictions using fixed thresholds \cite{zhang2021flexmatch}, this strategy remains imperfect. Furthermore, overconfidence and miscalibration of the model leading to incorrect predictions receive high confidence scores \cite{guo2017calibration, karimijafarbigloo2024leveraging}, allowing false pseudo-labels to persist. These errors degrade the learned knowledge from well-labeled data, further decreasing model performance. Moreover, using a fixed probability threshold is especially problematic in segmentation tasks. Semantic classes often vary in difficulty (i.e., difficult to segment in correct way), and harder classes are less likely to produce high-confidence predictions. This leads to biased pseudo-labeling, where easier or background classes dominate, and harder foreground regions are underrepresented. 

Motivated by previous challenges, we propose a dual-stream network architecture inspired by Knowledge Distillation (KD) frameworks \cite{li2024bridging}. Unlike traditional Student–Teacher paradigms, our method employs two parallel student networks, Subnet A and Subnet B, both based on 3D encoder–decoder architectures. To improve the quality and robustness of pseudo-labels, we introduce a Cross Consistency Enhancement (CCE) module, which combines Cross Pseudo Supervision (CPS) and Entropy-Filtered Supervision (EFS). The CPS mechanism allows each Subnet to generate pseudo-labels for the other, encouraging mutual refinement. EFS further filters noisy predictions by focusing supervision on low-entropy (i.e., high-confidence) voxels, improving pseudo-label reliability. {Our framework achieves robust bias correction and high-quality pseudo-labels by relying solely on inter-network interactions, eliminating the need for auxiliary teachers or additional correction networks.} In addition, we enforce consistency regularization by minimizing the Mean Squared Error (MSE) between the predictions of the two Subnets. This encourages consisetent outputs across different architectures and input perturbations, which improves generalization. To further mitigate the negative impact of noisy pseudo-labels, we incorporate uncertainty-aware training. Specifically, we assign dynamic voxel-wise weights to the pseudo-supervised loss based on prediction uncertainty (i.e., voxels with higher confidence receive greater weights, and vice versa). Finally, we integrate contrastive learning (CL) to improve feature-level discrimination. Uncertain voxel features are encouraged to align with class-specific prototypes of confidently predicted voxels, while maximizing inter-class separability in the feature space. Overall, this study introduces a novel dual framework in 3D medical image segmentation. {The key innovation lies in the prototype-guided design of label noise self-identification. This design identifies errors by dynamically comparing voxel features to class prototypes and exploiting the inherent anatomical consistency of medical imaging.} As shown in Figure \ref{fig:Visual}, the proposed method outperforms the existing SOTA methods in various data sets. Our contributions can be summarized as follows.


\begin{enumerate}
     \item We propose a novel CCE module that synergistically integrates Cross Pseudo Supervision and Entropy-Filtered Supervision to effectively suppress noisy pseudo-labels in medical image segmentation.
    \item We design an uncertainty-aware weighting mechanism that dynamically adjusts the contribution of pseudo-labels during training, effectively down-weighting unreliable voxels and reducing the negative impact of noisy supervision.
    \item We incorporate a prototype-based contrastive learning strategy to align uncertain voxel features with confidently predicted class prototypes, while maximizing the separation between different class features in the embedding space.
\end{enumerate}
The rest of this paper is organized as follows. In Section \ref{Related works}, we briefly introduce the related work on semi-supervised segmentation. In Section \ref{PROPOSED METHOD}, we present the proposed semi-supervised approaches. In Section \ref{Experiments}, we report and analyze the results of the experiment. In Section \ref{CONCLUSION}, we provide a summary of this study and outline future directions.

\section{Related works}\label{Related works}
\subsection{Semi-Supervised segmentation with pseudo labels}

{\textbf{Cross pseudo supervision.} CPS methods typically employ dual-network frameworks where predictions from one network guide another. However, it introduces complexity through auxiliary components. For instance, in \cite{wu2019mutual}, they propose a complementary correction network to refine pseudo-labels, and in \cite{mendel2020semi}, they introduced transformation rules to judge segmentation-image alignment. However, these approaches require complex loss functions (i.e., high computation overhead). In contrast, our framework eliminates auxiliary networks by leveraging direct inter-network interactions for bias correction, reducing computational overhead.}

{\textbf{Uncertainty estimation.} Uncertainty estimation is related to pseudo-label refinement. For example, Uncertainty-Aware Mean Teacher (UA-MT) \cite{yu2019uncertainty} employed Monte Carlo Dropout (MCD) to compute voxel-wise predictive entropy, while MC-Net \cite{wu2021semi} simplified this process by introducing an auxiliary classifier. Specifically, it designs two slightly different decoders to approximate epistemic uncertainty. Unlike UA-MT, which uses a weight-averaged teacher model for uncertainty measurement, Ambiguity-Consistent Mean Teacher (AC-MT) \cite{xu2023ambiguity} used a student model to reflect its current uncertainty. This is achieved by performing stochastic forward passes with the dropout layers enabled, which can be interpreted as Monte-Carlo samples of the posterior distribution of the model weights. Unlike these studies, the proposed model integrates uncertainty estimation with prototype guidance, enabling efficient noise identification without costly MCD.}

{\textbf{Prototypes based.} Prototypes have demonstrated robustness in handling label noise \cite{han2019deep} and domain shift \cite{zhang2021prototypical}. For example, SS-Net \cite{wu2022exploring} disentangled feature manifolds using prototypes. In \cite{wu2023exploring}, they propose a two-stage framework that leverages pseudo-labels to optimize feature spaces through feature representation learning. Mutual Learning with Robust Prototype Learning (ML-RPL) \cite{su2024mutual} computes class prototypes from pseudo-labels for mutual learning. In \cite{han2022effective}, they generate pseudo-labels by calculating the distances between feature vectors of unlabeled images and class representations derived from labeled data. Unlike others, we design a prototype-guided label noise self-identification to dynamically identify errors by comparing voxel features to class prototypes.}

{\textbf{Other approaches.} Recent works combine multiple strategies to improve reliability. For example, in \cite{yao2022enhancing}, they propose a confidence-aware cross-pseudo supervision network to improve pseudo-label quality by calculating pixel-wise confidence using Kullback–Leibler (KL) divergence between predictions of original and perturbed images. Furthermore, in \cite{aparco2025consensus}, they further proposed an iterative meta-pseudo-labeling approach based on a teacher-student framework. Unlike \cite{10980890}, which relies solely on uncertainty weighting, our approach integrates CPS and entropy filtering through a novel CCE module. Furthermore, we incorporate CL to align pixel features with class prototypes, enhancing intra-class compactness and inter-class separation in the feature space.}

\subsection{Semi-Supervised segmentation with consistency learning}
Consistency learning has emerged as a foundation in semi-supervised medical image segmentation, using the principle that the predictions of the model remain consistent under different perturbations of the same input. For example, in \cite{sajjadi2016regularization}, the authors introduced the \text{$\Pi$-Model}, which applies random augmentations to both labeled and unlabeled samples, requiring consistent outputs from the model when the same sample propagates through it under varying perturbations. In \cite{laine2016temporal}, they proposed Temporal Ensembling, combining predictions from multiple models trained at different time points using Exponential Moving Average (EMA). However, maintaining EMA predictions during training is computationally expensive. To address this, in \cite{tarvainen2017mean}, the authors developed the mean-teacher model, where a teacher model, updated with the EMA weights of the student model, guides the student through consistency constraints. In \cite{zeng2021reciprocal}, the authors incorporated feedback from the performance of the student model in labeled data to refine the updates of the teacher model through gradient descent, enhancing the adaptability of the consistency framework. In \cite{wu2024role}, a bidirectional self-training semi-supervised framework based on role exchange is proposed. This framework effectively improves the performance of complex medical image segmentation by dynamically switching the roles of the teacher and student networks. To enforce consistent outputs across different network architectures, we introduce a consistency regularization term based on the MSE between predictions of Subnet A and Subnet B. Furthermore, to mitigate the impact of noisy pseudo-labels, we adopt a dynamic uncertainty-aware weighting strategy, where each voxel contribution to the pseudo-supervised loss is adaptively scaled based on its estimated uncertainty—voxels with higher confidence receive greater weight.


\section{Proposed method} \label{PROPOSED METHOD}

\begin{figure*}
    \centering
    \includegraphics[width=0.9\linewidth]{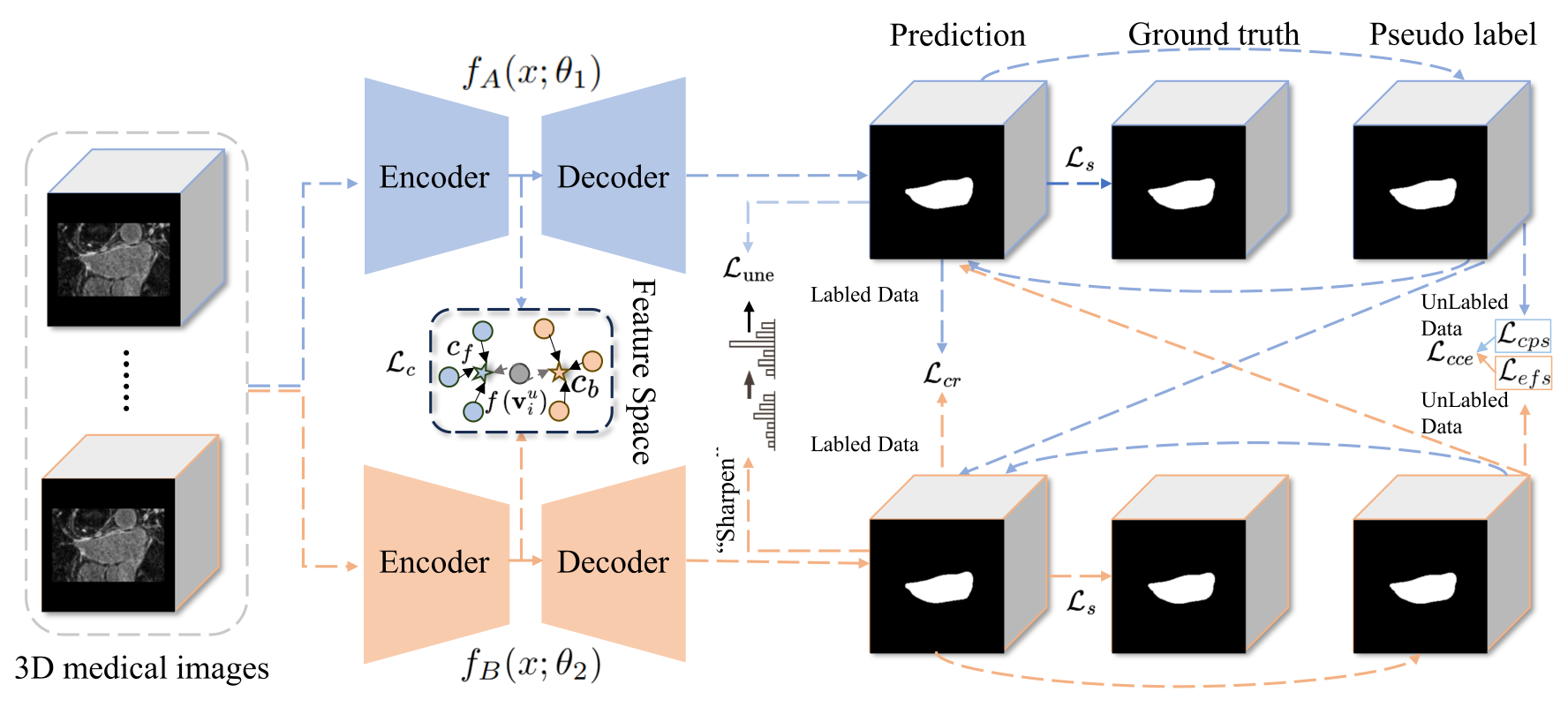}
    \caption{{Pipeline of semi-supervised segmentation framework. Both labeled and unlabeled 3D medical images are simultaneously fed into two 3D encoder-decoder networks to enhance segmentation performance: a ResNet34-based subnet \( f_A \) and a VNet-based subnet \( f_B \). The output features from both subnets are used to compute contrastive loss \(\mathcal{L}_c\) by aligning foreground and background prototypes in the feature space. The predicted segmentation maps are supervised using labeled data via a combined cross-entropy and Dice loss (\( \mathcal{L}_{s} \)). A consistency regularization loss (\( \mathcal{L}_{cr} \)) is applied on labeled data to encourage prediction stability, while an uncertainty estimation loss (\( \mathcal{L}_{une} \)) is applied on unlabeled data to suppress unreliable pseudo labels. Pseudo labels are generated from the model predictions and further refined using Cross Pseudo Supervision loss (\( \mathcal{L}_{cps} \)) and Entropy-Filtered Supervision loss (\( \mathcal{L}_{efs} \)) on unlabeled data.}}
    \label{fig:modelall}
\end{figure*}

\begin{figure}
    \centering
    \includegraphics[width=0.7\linewidth]{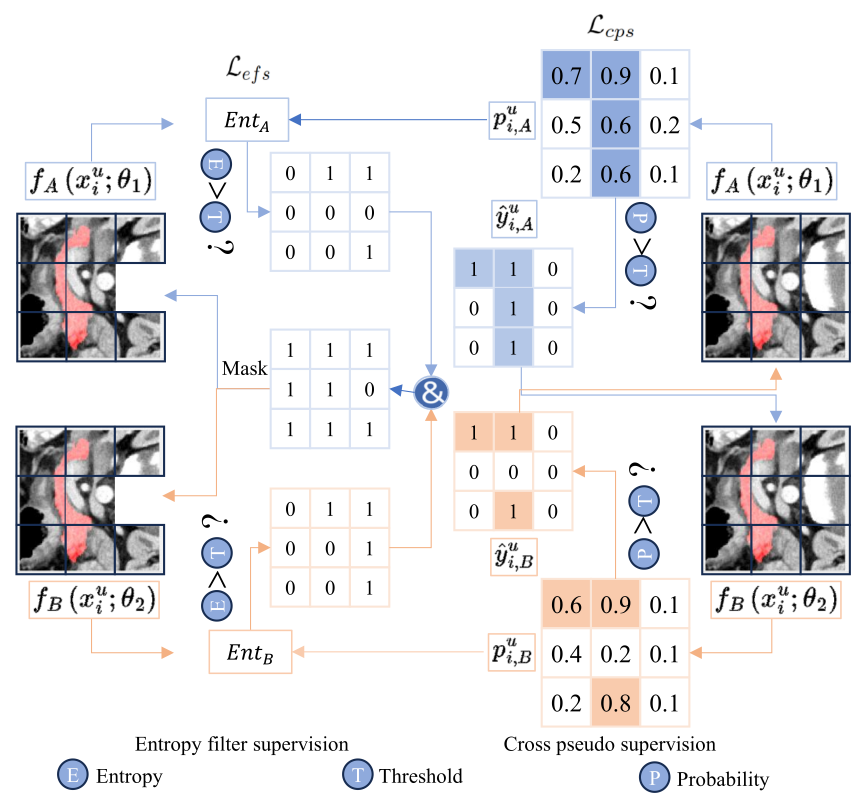}
    \caption{{The process of CCE module, which combines Cross Pseudo Supervision (CPS) with Entropy-Filtered Supervision (EFS) to improve the reliability of pseudo-labeling. In CPS, the predictions from one model serve as pseudo-labels for the other, where predictions exceeding a confidence threshold are assigned as 1 and others as 0. These pseudo-labels are then used to supervise the other model. In EFS, entropy maps are computed for each pixel’s predicted probability. Pixels with entropy exceeding a threshold in both models are masked out during training, ensuring that only low-uncertainty regions contribute to learning.}}
    \label{fig:modelCCE}
\end{figure}



\noindent\textbf{General framework.} Figure \ref{fig:modelall} presents the overall framework of our method. Basically, our model has two parallel 3D encoder-decoder subnetworks: Subnet A ($ f_A(x; \theta_1)$), and Subnet B ($f_B(x; \theta_2)$). Specifically, Subnet A and Subnet B are based on ResNet34- and V-Net based architectures, respectively. The proposed model leverages a combination of annotated samples $
D_l = \left\{ \left( x_i^l, y_i^l \right) \right\}_{i=1}^{N_l}$ and a larger set of unlabeled data $
D_u = \left\{ x_i^u \right\}_{i=1}^{N_u}$ to improve the performance. In addition, the consistency regularization (CR), uncertainty estimation (UE), and Prototype guided contrastive learning (PGL) are introduced to maximize the performance of the Subnet A and Subnet B. 

Overall, the total loss function is derived from three parts: 1) a supervised segmentation term $\mathcal{L}_s$, 2) a contrastive term $\mathcal{L}_c$ for self-supervised learning and 3) a pseudo-labeling term $\mathcal{L}_u$ for semi-supervised training.
\begin{equation}
\mathcal{L} = \underbrace{\mathcal{L}_s^{A} + \mathcal{L}_s^{B}}_{\textit{labelled}} 
+  \underbrace{\lambda_c \mathcal{L}_{c}}_{\textit{self}} 
+  \underbrace{\mathcal{L}_u^{A} + \mathcal{L}_u^{B}}_{\textit{semi}}.
\label{eq:overall_loss}
\end{equation}

For $\mathcal{L}_s$, the standard Cross-Entropy (CE) loss and Dice loss are used to optimize the model. Furthermore, a CR loss is introduced to improve the prediction consistency between Subnet A and Subnet B. The total supervised loss is defined as:
\begin{equation}
\mathcal{L}_s = \mathcal{L}_{ce}(f(x_i^l; \theta), y_i^l) + \text{Dice}(\hat{p}_i^l, y_i^l) + \alpha \cdot \mathcal{L}_{cr},
\label{eq:supervised_loss}
\end{equation}
where $y_i^l$ represents the mask label (ground truth) for the image labeled $i$, \(f(x_i^l; \theta)\) denotes the logits (model outputs before softmax activation), $\hat{p}_i^l$ represents the softmax outputs, and the hyperparameter $\alpha$ is to balance the supervised and consistency regularization loss.

For the unlabeled data, we combine Cross Pseudo Supervision (CPS) and an Entropy-Filtered Supervision (EFS) as the CCE module. Additionally, we incorporate an uncertainty estimation mechanism to enhance the learning process.
\begin{equation}
\mathcal{L}_u =  \mathcal{L}_{cce} +\mathcal{L}_{une} .
\label{eq:unsupervised_loss}
\end{equation}

We elaborate these components as follows.

\subsection{Consistency regularization}
We introduce a CR module to refine the predictions of the proposed model. This module helps the model identify regions where Subnets A and B provide inconsistent predictions, despite each Subnet high prediction confidence (i.e., potential mispredictions). The goal is to rectify these discrepancies. First, the pseudo-label \(\hat{y}_{i,B}^l\) for the Subnet B network is obtained by applying the argmax function to the softmax output:
\begin{equation}
\hat{y}_{i,B}^l = \arg\max (\sigma(f_B(x_i^l; \theta_2)))
\end{equation}
\(\hat{y}_{i,A}^l\) count in the same way. $\sigma$(·) indicates the softmax function.
This formula defines a binary mask $\mathcal{M}$ as follows.
\begin{equation}\footnotesize
\label{lambdaM}
    \mathcal{M} = \mathbb{I}\Big((\hat{y}_{i,B}^l = 1 \land f_B(x_i^l; \theta_2
) \geq \lambda) 
\oplus 
(\hat{y}_{i,A}^l = 1 \land f_A(x_i^l; \theta_1) \geq  \lambda
\Big),
\end{equation}
where \( \mathbb{I}(\cdot) \) is the indicator function that returns 1 if and only if one of the networks predicts the target class and its prediction exceeds the threshold. Otherwise, the indicator is 0. 
$\oplus$ represents the logical OR between the two conditions. $\lambda$ represents threshold.

{Finally, the prediction consistency from many models of the same input can be improved after minimizing the mean squared error as follows:}
\begin{equation}\label{EQ:6}
    \mathcal{L}_{cr} = \frac{  \|\mathcal{M} \cdot(f(x_i^l; \theta) - {y}_i^l)\|^2_2}{ \mathcal{M} + \epsilon},
\end{equation}
where $\epsilon$ is a small constant added for numerical stability to prevent division by zero.


\subsection{CCE module}

\subsubsection{Cross pseudo supervision}
In CPS, as shown in Figure \ref{fig:modelCCE}, the predictions from one model serve as pseudo-labels for the other. Specifically, if a model’s prediction exceeds a confidence threshold (e.g., 0.6), it is assigned as 1; otherwise, it is set to 0. These pseudo-labels are then used to supervise the other model. For each unlabeled data, the network generates a pseudo-label based on its predictions as follows:
\begin{equation}
    \hat{y}_i^u = \text{argmax}_c\hat{p}_i^u(c),
\end{equation}
where \( \hat{p}_i^u(c) \) represents the softmax output for class c. 

The CPS loss operates bidirectionally: one direction propagates from 
\( f_A(x_i^u; \theta_1) \) to \( f_B(x_i^u; \theta_2) \), and the other goes in reverse. Specifically, the pixel-wise one-hot label map \( \hat{y}_A^u \), generated by the network \( f_A(x_i^u; \theta_1) \), is used to supervise the pixel-wise confidence map \( {p}_{i,B}^u \) produced by the other network \( f_B(x_i^u; \theta_2) \) and vice versa. 


Similarly to feature or probability consistency, the proposed CPS enforces consistency between the two perturbed segmentation networks. This approach implicitly augments the training process by leveraging unlabeled data through the use of pseudo-labels. The CPS loss for the unlabeled data is defined as follows:
\begin{equation}
\label{losscps}
\mathcal{L}_{cps} = \mathcal{L}_{ce}(p_{i,B}^u, \hat{y}_{i,A}^u) + \mathcal{L}_{ce}(p_{i,A}^u, \hat{y}_{i,B}^u),
\end{equation}
where $p_{i,A}^u$ and $p_{i,B}^u$ are predicted probability from the student and teacher models. \(\hat{y}_{i, B}^u\) and \(\hat{y}_{i, A}^u\) are pseudo-labels generated by the student and teacher model. This framework enables the network to learn from stable pseudo-labels, improving the overall performance on both labeled and unlabeled data.

\subsubsection{Entropy-filtered supervision}

In EFS, as shown in Figure \ref{fig:modelCCE},  we compute the entropy of each pixel predicted probability. If the entropy value exceeds a certain threshold , the pixel is considered uncertain (1); otherwise, it is regarded as reliable (0). The entropy maps from both models are then intersected to build a mask, which is applied to the pseudo-labels. Pixels with entropy values exceeding this threshold in both models are masked out (ignored) during training, ensuring that only low-uncertainty regions contribute to learning. 

Before optimizing the unsupervised loss (UL), pixel-level entropy filtering is used to reduce the unreliable pseudo-labels. Specifically, the proposed UL is based on entropy minimization. This approach focuses on reducing the prediction uncertainty provided by the model and excludes regions with high uncertainty, preventing the model from being misled. Specifically, the entropy of each pixel is used to represent the confidence of the prediction. Furthermore, the regions with the highest uncertainty are filtered out based on a predefined threshold and marked with a special "ignore" label. Finally, we optimize the CE loss function with an ignored index to reduce the impact of noise as follows:
\begin{equation}
Ent_{A(B)} = - \sum_{c=1}^{C} \hat{p}_i^u(c) \cdot \log_2 \left( \hat{p}_i^u(c) \right).
\end{equation}
where \(Ent_A(\cdot)_{A(B)}\) represents the entropy value of the Subnet A and Subnet B, respectively.

We define a mask for valid points based on a threshold \(\tau\), while ignoring high-entropy pixels:
\begin{equation}
    \hat{y}^u_{i} =
\begin{cases}
\text{ignore}, & \text{if } Ent_A > \tau_A \text{ or } Ent_B > \tau_B , \\
\arg\max_c \, \hat{p}_{i}^u(c), & \text{otherwise.}
\end{cases}
\end{equation}
{where the thresholds \( \tau_A \) and \( \tau_B \) are chosen as the \( \gamma^{\text{th}}\) percentile of the entropy values predicted by each model ($\gamma$ is a constant value).}

We calculate the EFS loss as follows:
\begin{equation}
\label{lossefs}
\mathcal{L}_{efs} = \frac{\sum_{i=1}^{B \times H \times W \times D} \mathcal{L}_{ce}(f(x; \theta), \mathbb{I}\left[ \hat{y}^u_i \neq \text{ignore} \right]) }{\sum_{i=1}^{B \times H \times W \times D} \mathbb{I}\left[ \hat{y}^u_i \neq \text{ignore} \right]},
\end{equation}
where B, H, W, and D denote the batch size, height, width, and depth of the voxel grid, respectively. This approach helps filter out noisy pseudo-labels, improving training stability and model performance.

By combining CPS loss and EFS loss, we derive the CCE loss, formulated as follows:
\begin{equation}
\label{losscce}
\mathcal{L}_{cce} = \mathcal{L}_{cps} + \mathcal{L}_{efs}.
\end{equation}

\subsection{Uncertainty estimation}
Due to label noise, pseudo supervision can be unreliable. A common strategy to alleviate this issue is to apply a confidence threshold; however, this approach is impractical for segmentation tasks as it results in a biased model that focuses more on easier classes and less on harder ones. To address this limitation, we leverage prediction uncertainty to refine pseudo-label supervision. Specifically, we estimate uncertainty via KL-divergence \cite{zhao2023rcps, zheng2021rectifying}, and adjust the contribution of pseudo-labels based on confidence derived from both labeled and unlabeled data. To enhance reliability, we reformulate the softmax output as $\sigma(p^u/T_p)$ using a temperature scaling parameter $T_p$, which makes the pseudo labels more decisive and reduces class overlap. This operation also reinforces the low-density separation assumption in SSL \cite{van2020survey}, promoting class-wise clustering and encouraging decision boundaries to lie in low-density regions. As a result, the model generates more confident and less ambiguous pseudo labels. The uncertainty loss is defined as:
\begin{equation}
\label{losstp}
\mathcal{L}_{p}(p_{i,A}^u, p_{i,B}^u) = \mathcal{L}_{ce}(p_{i,A}^u, \sigma(p_{i,B}^u / T_p)),
\end{equation}
\begin{equation}\footnotesize
\mathcal{L}_{une}(\hat p_{i,A}^u, \hat p_{i,B}^u) = e^{-\mathcal{D}_{KL}(\hat p_{i,A}^u, \hat p_{i,B}^u)} \mathcal{L}_{p}(\hat p_{i,A}^u, \hat p_{i,B}^u) + \mathcal{D}_{KL}(\hat p_{i,A}^u, \hat p_{i,B}^u),
\end{equation}
\begin{equation}
\mathcal{D}_{KL}(\hat p_{i,A}^u, \hat p_{i,B}^u) = \hat p_{i,B}^u \log \left(\frac{\hat p_{i,B}^u}{\hat p_{i,A}^u}\right),
\end{equation}
where, $\mathcal{D}_{KL}(\hat p_{i,A}^u, \hat p_{i,B}^u)$ represents the KL-divergence between the prediction and the pseudo label provided by the model, helping to reduce the influence of noisy labels by incorporating uncertainty information into the loss.  \(\hat p_{i,A}^u \) and \(\hat p_{i,B}^u \)
represent the softmax probability distributions output by the Subnet A and Subnet B, respectively.


\begin{algorithm}\footnotesize

\caption{{Training process of the propose model.}}
\textbf{Input:} Subnet $f_A$; Subnet $f_B$; Labeled dataset $\mathcal{D}_l$; Unlabeled dataset $\mathcal{D}_u$ \\
\textbf{Output:} Trained $f_A$ and $f_B$
\begin{algorithmic}
    \State Initialize $f_A$ and $f_B$ using same Kaiming initialization
    \While{not converged}
        \State {Compute $\mathcal{L}_s$ using Eq. \eqref{eq:supervised_loss}} \Comment{\textcolor{olive}{Supervised loss}}
       
        \State Compute $\mathcal{L}_{cps}$ using Eq. \eqref{losscps}
        \Comment{\textcolor{olive}{Cross pseudo supervision loss}}
        \State Compute  $\mathcal{L}_{efs}$ using Eq. \eqref{lossefs} \Comment{\textcolor{olive}{Entropy-filtered supervision loss}}
        \State $\mathcal{L}_{cce} = \mathcal{L}_{cps} + \mathcal{L}_{efs}$ \Comment{\textcolor{olive}{Total consistency loss}}
        \State Compute $\mathcal{L}_{une}$ using Eq. \eqref{losstp} \Comment{\textcolor{olive}{Uncertainty estimation loss}}
        \State $\mathcal{L}_u = \mathcal{L}_{cce} + \mathcal{L}_{une}$ \Comment{\textcolor{olive}{Total unsupervised loss}}
        \State Compute $c_f$ (foreground) and $c_b$ (background) using Eq. (\ref{EQ:cf})
        \State Compute  $\mathcal{L}_{c}$ using Eq. (\ref{EQ:contras}) \Comment{\textcolor{olive}{Contrastive loss}}
        \State Update $f_A$ and $f_B$ using total loss $\mathcal{L} = \mathcal{L}_s + \mathcal{L}_c + \mathcal{L}_u$;
    \EndWhile
\end{algorithmic}
\label{Alg.1a}
\end{algorithm}

\subsection{Prototype guided contrastive learning}

\noindent\textbf{Prototype.} 
To alleviate prediction ambiguity (e.g., misclassification risk) and enhance supervision in the feature space (e.g., improve confidence estimation), we introduce a PGL function that aligns uncertain voxels with their associated class prototypes. Specifically, the model evaluates the certainty of each voxel and separates them into reliable and unreliable groups. Class-wise prototypes are then computed by averaging features from the reliable set, providing discriminative guidance that facilitates more accurate semantic differentiation.

First, the average feature vectors (prototypes) for reliable voxels are computed separately for foreground and background classes. {We define voxel-wise reliability by jointly considering two criteria: (1) prediction agreement between the two models \( f_A \) and \( f_B \) and (2) low predictive entropy indicating high confidence. Specifically, we compute the voxel-wise entropy \( Ent_A(\mathbf{v}_i) \) and \( Ent_B(\mathbf{v}_i) \) from the softmax outputs of each model.  
Furthermore, a confidence threshold \( \tau \) is set as the \( \gamma^{\text{th}} \) percentile of all entropy values in a batch.  
A binary reliability mask \( M_i \in \{0, 1\} \) is constructed as:}
{\begin{equation}
M_i =
\begin{cases}
1, & \text{if } (\arg\max\hat{p}_{i, A}^u = \arg\max\hat{p}_{i,B}^u) \text{ and } (Ent_A(\mathbf{v}_i) < \tau_A \text{ and } Ent_B(\mathbf{v}_i) < \tau_B) \\
0, & \text{otherwise}
\end{cases}
\end{equation}}

{Accordingly, we define the reliable set and uncertain set as:}
{\begin{equation}
S^r = \{ \mathbf{v}_i \mid M_i = 1 \}, \quad S^u = \{ \mathbf{v}_i \mid M_i = 0 \}
\end{equation}}

{The reliable set \( S^r \) is further divided into class-specific subsets:}
{\begin{equation}
S_f^r = \{ \mathbf{v}_i \in S^r \mid \hat{y}_i = 1 \}, \quad S_b^r = \{ \mathbf{v}_i \in S^r \mid \hat{y}_i = 0 \}
\end{equation}}
{where \( \hat{y}_i = \arg\max(f_A(\mathbf{v}_i)) \). These subsets are used to compute the class-wise feature prototypes.}

The remaining uncertain voxels in \( S^u \) are excluded from supervised loss and used to construct contrastive pairs for prototype-based feature alignment.
The class prototypes are then computed as the mean feature embedding within each reliable subset, formulated as:

\begin{equation}\label{EQ:cf}
    c_f = \frac{1}{|S_f^r|} \sum_{\mathbf{v}_i^r \in S_f^r} f(\mathbf{v}_i^r), \quad
    c_b = \frac{1}{|S_b^r|} \sum_{\mathbf{v}_i^r \in S_b^r} f(\mathbf{v}_i^r),
\end{equation}

where \( c_f \) and \( c_b \) denote the prototypes of the foreground and background classes, respectively. Here, \( f(\mathbf{v}_i^r) \) represents the feature embedding of the reliable voxel \( \mathbf{v}_i^r \).

\begin{table}[!ht]
     \caption{Performance metrics for segmentation techniques employed on the Left Atrium dataset. \textbf{Bold} means the best result. Model complexity is reported in terms of the number of parameters (Para.) and multiply-accumulate operations (MACs), measured during inference. MACs are computed per 3D image with input dimensions H × W × D (height × width × depth).}
    \label{tab:LAcomparison}
    \adjustbox{width=1\linewidth}{
    \begin{tabular}{lcccccccc}
        \toprule
    \multirow{2}{*}{Method} & \multicolumn{2}{c}{Scans used} & \multicolumn{4}{c}{Metrics}  &  \multicolumn{2}{c}{Complexity}\\ \cmidrule(lr){2-3} \cmidrule(l){4-9}
     & Labeled & Unlabeled & Dice(\%) $\uparrow$ & Jaccard(\%) $\uparrow$ & 95HD(voxel) $\downarrow$ & ASD(voxel) $\downarrow$ & Para.(M)&MACs(G) \\
    \midrule
        V-Net \cite{milletari2016v} (SupOnly) & 8 (10\%) & 0 &51.76 ± 7.38&41.14 ± 6.57&31.44 ± 3.43&7.88 ± 1.43&9.44 &47.10 \\
        V-Net \cite{milletari2016v} (SupOnly) & 16 (20\%) & 0 &77.48 ± 4.94&65.27 ± 5.12&22.30 ± 3.58&6.26 ± 0.91&9.44 &47.10 \\
        V-Net \cite{milletari2016v} (SupOnly) & 80 (All) & 0 &90.62 ± 0.61&83.04 ± 1.01&6.27 ± 1.28&1.81 ± 0.28&9.44&47.10 \\
    \midrule
       UA-MT \cite{yu2019uncertainty} (MICCAI’19)  &  &  & 87.10 ± 1.09 & 77.47 ± 1.64 & 14.22 ± 2.73 & 3.69 ± 0.46 &18.88&94.80\\
SASSNet \cite{li2020shape} (MICCAI’20)  & & & 84.39 ± 2.49 & 74.36 ± 3.18 & 11.96 ± 2.06 & 3.22 ± 0.37&20.46&15.41 \\
DTC \cite{luo2021semi} (AAAI’21)   & & & 87.96 ± 1.34 & 78.96 ± 1.91 & 6.93 ± 0.10 & 2.07 ± 0.27&9.44&47.10 \\
MC-Net \cite{wu2021semi} (MICCAI’21)  & & & 86.80 ± 1.25 & 77.08 ± 1.86 & 13.16 ± 2.73 & 3.77 ± 0.68&12.35&95.32 \\
SS-Net \cite{wu2022exploring} (MICCAI’22)  & 8 (10\%) & 72(90\%)  & 88.82 ± 0.72 & 80.03 ± 1.16 & 7.34 ± 0.92 & 2.04 ± 0.21&9.45&47.10 \\
MC-Net+ \cite{wu2022mutual} (MedIA’22) & & & 87.85 ± 1.13 & 78.68 ± 1.72 & 7.75 ± 1.03 & 1.96 ± 0.23&15.25&143.37 \\
MCF \cite{wang2023mcf} (CVPR’23)   & & & 86.10 ± 1.97 & 76.54 ± 2.73 & 7.69 ± 1.10 & 2.67 ± 0.38 &27.30&126.95\\
BCP \cite{bai2023bidirectional} (CVPR’23) &&& 89.45 ± 0.58&81.02 ± 0.94&8.62 ± 1.75&2.79 ± 0.37&28.32&141.90\\
LUSEG \cite{karimijafarbigloo2024leveraging} (ISBI’24)  & & & 87.17 ± 1.94 & 77.92 ± 2.70 & 7.46 ± 1.17 & 2.54 ± 0.42&27.30 &126.95\\
ML-RPL \cite{su2024mutual}(MedIA’24)&&& 87.12 ± 1.09&77.49 ± 1.63&9.47 ± 1.43&2.93 ± 0.36&15.25&143.37\\
\rowcolor{gray!30} \textbf{Ours} & & &\textbf{89.95 ± 0.58} & \textbf{81.81 ± 0.95} & \textbf{5.73 ± 0.63} & \textbf{1.79 ± 0.24}&27.30&126.95 \\
        \midrule
        UA-MT \cite{yu2019uncertainty} (MICCAI’19)  &  &&90.17 ± 0.59&82.21 ± 0.97& 8.15 ± 1.95& 2.36 ± 0.42&18.88&94.80\\
        SASSNet \cite{li2020shape} (MICCAI’20) & & & 90.33 ± 0.55 &82.45 ± 0.91 &5.94 ± 0.56 &1.86 ± 0.15&20.46&15.41\\
        DTC \cite{luo2021semi} (AAAI’21) & & &90.02 ± 0.76&  82.03 ± 1.22& 6.02 ± 0.71& 1.75 ± 0.19&9.44&47.10  \\
        MC-Net \cite{wu2021semi} (MICCAI’21) & & &91.21 ± 0.50& \textbf{83.92 ± 0.83}& 5.56 ± 0.62& \textbf{1.48 ± 0.14} &12.35&95.32\\
        SS-Net \cite{wu2022exploring} (MICCAI’22) & 16 (20\%) & 64(80\%)&90.55 ± 0.52 &82.81 ± 0.86&5.83 ± 0.65& 1.66 ± 0.15&9.45&47.10 \\
        MC-Net+ \cite{wu2022mutual} (MedIA’22)& & &91.14 ± 0.49& 83.80 ± 0.81& 5.52 ± 0.60 & 1.50 ± 0.15&15.25&143.37\\
        MCF \cite{wang2023mcf} (CVPR’23) & & &90.05 ± 0.89& 82.12 ± 1.37& 5.99 ± 0.81 &1.72 ± 0.22&27.30&126.95 \\
        BCP \cite{bai2023bidirectional} (CVPR’23)  &&&  \textbf{91.61 ± 0.44}&84.58 ± 0.74&\textbf{4.97 ± 0.59}&1.78 ± 0.19 &28.32&141.90\\
        LUSEG \cite{karimijafarbigloo2024leveraging} (ISBI’24)& & &90.18 ± 0.75& 82.28 ± 1.20& 5.77 ± 0.69& 1.78 ± 0.20 &27.30 &126.95\\
        ML-RPL \cite{su2024mutual}(MedIA’24)&&&89.28 ± 0.96&80.90 ± 1.50&12.01 ± 3.02&3.49 ± 0.78&15.25&143.37\\
        \rowcolor{gray!30} \textbf{Ours} & & &90.90 ± 0.54 &83.41 ± 0.90& 5.28 ± 0.58& 1.63 ± 0.15&27.30&126.95  \\
        \bottomrule
    \end{tabular}
    }
\end{table}

\begin{table}
    \centering
    \caption{Summary of performance metrics for segmentation techniques used on the Pancreas dataset. Model complexity is reported in terms of the size of the parameters (Para.) and multiply-accumulate operations (MACs), measured during inference. MACs are computed per 3D image with input dimensions H × W × D (height × width × depth).}
    \label{tab:PANCREAScomparison}
    \adjustbox{width=1\linewidth}{
    \begin{tabular}{lcccccccc}
        \toprule
    \multirow{2}{*}{Method} & \multicolumn{2}{c}{Scans used} & \multicolumn{4}{c}{Metrics}&  \multicolumn{2}{c}{Complexity} \\ \cmidrule(lr){2-3} \cmidrule(l){4-9}
     & Labeled & Unlabeled & Dice(\%) $\uparrow$ & Jaccard(\%) $\uparrow$ & 95HD(voxel) $\downarrow$ & ASD(voxel)$\downarrow$& Para.(M)&MACs(G) \\
    \midrule
        V-Net \cite{milletari2016v} (SupOnly) & 6 (10\%) & 0 &60.77 ± 5.80&46.37 ± 5.01&22.08 ± 4.98&6.67 ± 1.46&9.44 &41.53\\
        V-Net \cite{milletari2016v} (SupOnly) & 12 (20\%) & 0 &69.21 ± 4.18&55.34 ± 4.21&12.76 ± 2.15&2.49 ± 0.69&9.44 &41.53 \\
        V-Net \cite{milletari2016v} (SupOnly)  & 62 (All) & 0 &80.30 ± 1.82&67.80 ± 2.40&6.45 ± 0.77&1.62 ± 0.11&9.44 &41.53\\
    \midrule
        UA-MT \cite{yu2019uncertainty} (MICCAI’19) & & & 69.92 ± 3.10 & 55.31 ± 3.31 & 23.06 ± 2.86 & 7.65 ± 0.88&18.88 &83.06 \\
SASSNet \cite{li2020shape} (MICCAI’20)  & & & 69.34 ± 4.18 & 55.64 ± 4.08 & 14.65 ± 2.31 & 3.69 ± 0.57&20.46&13.59 \\
DTC  \cite{luo2021semi} (AAAI’21)     & & & 69.45 ± 3.18 & 54.88 ± 3.49 & 14.71 ± 1.87 & 2.99 ± 0.46&9.44&41.53 \\
MC-Net \cite{wu2021semi} (MICCAI’21)   & & & 68.55 ± 3.99 & 54.52 ± 3.96 & 14.16 ± 2.17 & 2.37 ± 0.23&12.35&84.04 \\
SS-Net \cite{wu2022exploring} (MICCAI’22)   & 6 (10\%) & 56(90\%)& 66.48 ± 4.50 & 52.72 ± 4.37 & 24.23 ± 6.78 & \textbf{1.92 ± 0.24} &9.45&41.53\\
MC-Net+ \cite{wu2022mutual} (MedIA’22)  & & & 68.07 ± 4.05 & 53.88 ± 3.80 & 23.86 ± 4.33 & 5.88 ± 1.40&15.25&126.4 \\
MCF   \cite{wang2023mcf} (CVPR’23)   & & & 70.12 ± 3.65 & 56.04 ± 3.70 & 13.16 ± 2.37 & 3.77 ± 0.58 &27.30&111.92\\
 BCP \cite{bai2023bidirectional} (CVPR’23)  &&& 57.35 ± 4.15&42.34 ± 3.73&29.85 ± 3.56&8.19 ± 0.99 &28.32&141.90\\
LUSEG  \cite{karimijafarbigloo2024leveraging} (ISBI’24)  & & & 68.33 ± 2.70 & 53.64 ± 3.03 & 19.30 ± 1.83 & 6.04 ± 0.50&27.30&111.92 \\
 ML-RPL \cite{su2024mutual}(MedIA’24)&&&70.37 ± 2.97&55.78 ± 3.27&17.30 ± 2.73&4.75 ± 0.72&15.25&143.37\\
        \rowcolor{gray!30} \textbf{Ours} & &  &\textbf{71.83 ± 4.15}&\textbf{57.87 ± 3.97}&\textbf{11.26 ± 3.57}&2.75 ± 0.72&27.30&111.92 \\
        \midrule
        UA-MT \cite{yu2019uncertainty} (MICCAI’19) & &  &77.00 ± 1.96&  63.41 ± 2.56& 16.65 ± 3.43&  4.76 ± 0.82&18.88 &83.06\\
        SASSNet \cite{li2020shape} (MICCAI’20) & & &75.80 ± 2.76&  62.42 ± 3.15& 12.18 ± 2.61&  2.41 ± 0.31&20.46&13.59\\
        DTC \cite{luo2021semi} (AAAI’21) & & &74.77 ± 2.98& 61.34 ± 3.46& 9.30 ± 1.75& 1.77 ± 0.18&9.44&41.53  \\
        MC-Net \cite{wu2021semi} (MICCAI’21) & & &75.50 ± 2.43&  61.79 ± 2.95& 14.98 ± 2.70&  3.79 ± 0.56 &12.35&84.04\\
        SS-Net \cite{wu2022exploring} (MICCAI’22) & 12 (20\%) & 50(80\%)&69.14 ± 4.12& 55.52 ± 4.28& 22.72 ± 6.17&  \textbf{1.54 ± 0.14}&9.45&41.53\\
        MC-Net+ \cite{wu2022mutual} (MedIA’22)& & &76.07 ± 2.66& 62.72 ± 3.16& 10.39 ± 1.70&   2.48 ± 0.31&15.25&126.4\\
        MCF \cite{wang2023mcf} (CVPR’23) & & &73.47 ± 3.47&60.10 ± 3.75& 9.38 ± 1.48& 2.28 ± 0.26&27.30&111.92 \\
         BCP \cite{bai2023bidirectional} (CVPR’23)  && & 73.91 ± 2.28&59.58 ± 2.69&12.32 ± 2.79&2.65 ± 0.24&28.32&141.90   \\
        LUSEG \cite{karimijafarbigloo2024leveraging} (ISBI’24)& & &74.11 ± 3.17&60.68 ± 3.60& 9.68 ± 1.63& 2.09 ± 0.31&27.30&111.92 \\
         ML-RPL \cite{su2024mutual}(MedIA’24)&&&76.75 ± 2.47&63.46 ± 2.97&15.65 ± 5.29&1.62 ± 0.24&15.25&143.37\\
        \rowcolor{gray!30} \textbf{Ours} & & &\textbf{77.07 ± 3.86}& \textbf{64.35 ± 3.93}& \textbf{9.29 ± 1.79}&1.73 ± 0.14&27.30&111.92\\
        \bottomrule
    \end{tabular}
    }
\end{table}

\begin{table}[]
    \centering
    \caption{Performance metrics for segmentation techniques used on Brats-2019 dataset. Model complexity is reported in terms of the size of the parameters (Para.) and multiply-accumulate operations (MACs), measured during inference. MACs are computed per 3D image with input dimensions H × W × D (height × width × depth).}
    \label{tab:BRAINcomparison}
    \adjustbox{width=1\linewidth}{
    \begin{tabular}{lcccccccc}
        \toprule
    \multirow{2}{*}{Method} & \multicolumn{2}{c}{Scans used} & \multicolumn{4}{c}{Metrics} &  \multicolumn{2}{c}{Complexity}\\ \cmidrule(lr){2-3} \cmidrule(l){4-9}
     & Labeled & Unlabeled & Dice(\%) $\uparrow$ & Jaccard(\%) $\uparrow$ & 95HD(voxel)$\downarrow$ & ASD(voxel)$\downarrow$& Para.(M)&MACs(G) \\
    \midrule
        V-Net \cite{milletari2016v} (SupOnly) & 25 (10\%) & 0 & 74.43 ± 2.81 & 61.86 ± 3.12 &37.11 ± 1.66 &  2.79 ± 0.54&9.44 &41.53 \\
        V-Net \cite{milletari2016v} (SupOnly) & 50 (20\%) & 0 & 80.16 ± 1.93 & 71.55 ± 2.39 &  22.68 ± 1.54& 3.43 ± 0.43&9.44 &41.53\\
        V-Net \cite{milletari2016v} (SupOnly)& 250 (All) & 0  & 85.93 ± 1.24 & 76.81 ± 1.76 &  9.85 ± 2.12& 1.93 ± 0.54&9.44 &41.53\\
    \midrule
        UA-MT \cite{yu2019uncertainty} (MICCAI’19)  & & & 80.72 ± 1.66 & 70.30 ± 2.14 & 11.76 ± 1.96 & 2.72 ± 0.72&18.88 &83.06\\
        DTC \cite{luo2021semi} (AAAI’21) & & & 81.75 ± 1.43 & 71.63 ± 1.93 & 15.73 ± 1.96 & 2.56 ± 0.62&9.44&41.53\\
        URPC \cite{luo2022semi} (MedIA’22)&  25 (10\%) & 225(90\%)& 82.59 ± 1.84 & 72.11 ± 2.31 & 13.88 ± 1.70 & 3.72 ± 0.58&5.88&61.43\\
        CPCL \cite{xu2022all} (JBHI’22) & & & 83.10 ± 1.77 & 73.23 ± 2.28& 11.74 ± 2.00& 1.99 ± 0.22&18.88&83.06 \\
        AC-MT \cite{xu2023ambiguity} (MedIA’23)& & & 83.77 ± 1.65 & 73.77 ± 2.09 & 11.37 ± 2.46 & \textbf{1.93 ± 0.23}&18.88&83.06 \\
         BCP \cite{bai2023bidirectional} (CVPR’23)  && & 75.18 ± 3.26&65.31 ± 3.31&21.59 ± 3.42&9.54 ± 2.18&28.32&141.90    \\
          ML-RPL \cite{su2024mutual}(MedIA’24)&&&82.58 ± 1.97&72.69 ± 2.38&11.34 ± 1.59&2.95 ± 0.48&15.25&143.37\\
        \rowcolor{gray!30} \textbf{Ours} & & & \textbf{84.84 ± 1.45} & \textbf{75.13 ± 1.94} & \textbf{10.01 ± 1.70} & 2.93 ± 0.54&27.30&111.92 \\
        \midrule
        UA-MT \cite{yu2019uncertainty} (MICCAI’19) & & & 83.12 ± 1.70& 73.01 ± 2.18 & 9.87 ± 1.94 & 2.30 ± 0.60&18.88 &83.06\\
        DTC \cite{luo2021semi} (AAAI’21) & & &83.43 ± 1.98 & 73.56 ± 2.35 & 14.77 ± 1.97 & 2.34 ± 0.62&9.44&41.53\\
        URPC \cite{luo2022semi} (MedIA’22)&50 (20\%) & 200(80\%)& 82.93 ± 1.75& 72.57 ± 2.23 & 15.93 ± 1.54 & 4.19 ± 0.51&5.88&61.43 \\
        CPCL \cite{xu2022all} (JBHI’22) & & & 83.48 ± 2.18 & 74.08 ± 2.49 & 9.53 ± 1.16 & 2.08 ± 0.30&18.88&83.06  \\
        AC-MT\cite{xu2023ambiguity}  (MedIA’23)  & & & 84.63 ± 1.91 &74.39 ± 2.33& 9.50 ± 1.63 & 2.11 ± 0.32&18.88&83.06   \\
        BCP \cite{bai2023bidirectional} (CVPR’23) && &78.99 ± 2.51&68.75 ± 2.83&11.87 ± 1.88&2.24 ± 0.37&28.32&141.90 \\
         ML-RPL \cite{su2024mutual}(MedIA’24)&&& 84.55 ± 1.63&75.02 ± 2.12&7.79 ± 1.03&1.81 ± 0.29&15.25&143.37\\
        \rowcolor{gray!30} \textbf{Ours} & & &\textbf{85.76 ± 1.41}& \textbf{76.51 ± 1.91}& \textbf{7.46 ± 1.09}& \textbf{1.99 ± 0.32} &27.30&111.92\\
        \bottomrule
    \end{tabular}
    }
\end{table}

\begin{table*}
\centering
\renewcommand{\arraystretch}{1.3}
\caption{{Ablation study with each component, the ratio of labeled data is set to 10\%.}}
\label{Ablation}
\small
\resizebox{\textwidth}{!}{
\begin{tabular}{c|c|c|c|c|c|c|c|c|c|c|c|c|c|c|c|c|c}
\hline
\multirow{2}{*}{\centering $CPS$}&\multirow{2}{*}{\centering $EFS$}&\multirow{2}{*}{\centering $CCE$} & \multirow{2}{*}{\centering $PGL$} & \multirow{2}{*}{\centering $UE$} & \multirow{2}{*}{\centering $CR$} & \multicolumn{4}{c|}{LA} & \multicolumn{4}{c|}{Pancreas-CT} & \multicolumn{4}{c}{Brats-2019} \\
\cline{7-18}
& & & & && DSC (\%)  & Jaccard (\%)  & HD95 (voxel)  & ASD (voxel)  & DSC (\%)  & Jaccard (\%)  & HD95 (voxel) & ASD (voxel) & DSC (\%)  & Jaccard (\%)  & HD95 (voxel) & ASD (voxel) \\
\hline 
\checkmark & & & & & &89.70 ± 0.56 & 81.41 ± 0.92 & 6.03 ± 0.59 & 1.91 ± 0.16  & 68.74 ± 4.03 & 54.76 ± 3.95 & 15.86 ± 3.76 & 2.53 ± 0.38 & 82.86 ± 1.78 & 72.79 ± 2.27 & 10.28 ± 1.70 & 2.70 ± 0.44 \\
& \checkmark & & & & & 86.66 ± 1.51 & 77.03 ± 2.18 & 9.86 ± 1.89 & 3.14 ± 0.41 & 67.76 ± 3.95 & 53.51 ± 3.87 & 20.66 ± 2.90 & 5.82 ± 0.82 & 82.20 ± 1.99 & 72.20 ± 2.41 & 11.18 ± 1.89 & 3.31 ± 0.58 \\
& & \checkmark & & & & 89.78 ± 0.58 & 81.55 ± 0.95 & 5.97 ± 0.60 & 1.86 ± 0.14 & 71.11 ± 3.46 & 57.11 ± 3.63 & 11.53 ± 1.86 & 2.40 ± 0.42 &  83.41 ± 1.73 & 73.49 ± 2.22 & 9.45 ± 1.44 & 2.44 ± 0.38 \\
 && & \checkmark &\checkmark &\checkmark &88.37 ± 1.23 & 79.56 ± 1.78 & 6.69 ± 0.96 & 2.33 ± 0.28 & 68.26 ± 3.54 & 54.04 ± 3.64 & 19.94 ± 3.99 & 5.39 ± 1.06 & 80.94 ± 2.08 & 70.63 ± 2.55 & 11.14 ± 1.73 & 2.96 ± 0.47
\\
& & \checkmark &\checkmark &\checkmark & &89.47 ± 0.61& 81.05 ± 0.99 & 6.23 ± 0.54 & 2.09 ± 0.20 &4.86 ± 2.31&2.75 ± 1.35  &58.99 ± 11.89  &\textbf{2.27 ± 2.50} & 79.03 ± 2.52 & 68.99 ± 2.96 & 10.23 ± 1.44 & 2.42 ± 0.40
 \\
&& \checkmark& &\checkmark &\checkmark &89.77 ± 0.58& 81.55 ± 0.95 & 6.25 ± 0.63 &2.05 ± 0.24  & 70.78 ± 4.07 & 57.27 ± 3.86 & 13.21 ± 3.43 & 2.50 ± 0.77 & 84.82 ± 1.45 & 75.10 ± 1.94 & 10.04 ± 1.71 & 2.95 ± 0.54
\\
 &&\checkmark & \checkmark & &\checkmark &89.93 ± 0.53 & 81.77 ± 0.87 & 5.77 ± 0.58 & 1.80 ± 0.14 & 67.71 ± 4.66 & 54.13 ± 4.29 & 14.21 ± 2.87 & 2.77 ± 0.86 & 83.34 ± 1.68 & 73.30 ± 2.18 & \textbf{8.98 ± 1.34} & \textbf{2.33 ± 0.36}\\
 && \checkmark & \checkmark &\checkmark &\checkmark&\textbf{89.95 ± 0.58} & \textbf{81.81 ± 0.95} & \textbf{5.73 ± 0.63} & \textbf{1.79 ± 0.24}&\textbf{71.83 ± 4.15}&\textbf{57.87 ± 3.97}&\textbf{11.26 ± 3.57}&2.75 ± 0.72& \textbf{84.84 ± 1.45} & \textbf{75.13 ± 1.94} & 10.01 ± 1.70 & 2.93 ± 0.54
\\
\hline
\end{tabular}}
\resizebox{0.8\textwidth}{!}{CCE: Cross Consistency Enhancement; PGL: Prototype Guided Contrastive Learning; UE: Uncertainty estimation; CR: Consistency regularization}
\end{table*}



\noindent\textbf{Contrastive loss.} To encourage more discriminative feature representations, the contrastive loss consists of three components: it pulls uncertain voxels toward their respective class prototypes and promotes separation between different classes. The initial formulation is:

\begin{equation}\label{EQ:contras}
    \mathcal{L}_{c} = \sum_{\mathbf{v}_i^u \in S_f^u} d\big(f(\mathbf{v}_i^u), c_f\big) + \sum_{\mathbf{v}_i^u \in S_b^u} d\big(f(\mathbf{v}_i^u), c_b\big) + d(c_f, c_b),
\end{equation}

where \( f(\mathbf{v}_i^u) \) denotes the feature embedding of an uncertain voxel \( \mathbf{v}_i^u \), and \( d(\cdot, \cdot) \) is a distance metric, e.g., Euclidean distance.

Algorithm \ref{Alg.1a} details the training procedure of the proposed approach.

\section{Experiments}
\label{Experiments}
\subsection{Datasets}
\noindent\textbf{Left Atrial (LA).} As described in \cite{xiong2021global}, the LA dataset includes 100 3D MR volumes with gadolinium contrast and manual annotations of the left atrium, sampled at an isotropic resolution of 0.625 mm³. Following \cite{wang2023mcf, karimijafarbigloo2024leveraging}, the images are normalized to the zero mean and unit variance. Training data are randomly cropped to 112×112×80, while inference uses a sliding window of the same size and a stride of 18×18×4.

\noindent\textbf{NIH Pancreas.} This dataset \cite{roth2015deeporgan} consists of 82 contrast-enhanced 3D CT scans with pancreas annotations. Each scan has a resolution of 512×512 pixels and 181–466 slices along the axial direction. Following \cite{wang2023mcf, karimijafarbigloo2024leveraging}, soft tissue intensity clipping is applied in the range of -120 to 240 Hounsfield Unit (HU), and pancreas regions are cropped with a 25-voxel padding. Training inputs are obtained by random cropping to 96×96×96, and a 16×16×16 sliding window is used in inference.

\noindent\textbf{BraTS-2019.} This dataset \cite{hdtd-5j88-20} consists of MRI scans from 335 glioma patients (259 HGG and 76 LGG), with four imaging modalities per subject: T1, T1Gd, T2, and FLAIR. Annotations are provided at the voxel level and verified by neuroradiologists. We follow the data split and preprocessing pipeline of \cite{xu2023ambiguity}, using 250 cases for training, 25 for validation, and 60 for testing. Random 96×96×96 patches are used for training, and testing is performed using a sliding window with a stride of 64×64×64.

\subsection{Implementation details}
Our experimental configuration follows the protocol in \cite{su2024mutual}, employing ResNet and V-Net as dual-stream backbones to enable fair comparison versus existing approaches. The network parameters are updated using stochastic gradient descent (SGD) with a momentum of 0.9 and a weight decay of $1\times10^{-4}$. The initial learning rate is 0.01 and decays tenfold every 2500 steps, for a total of 6000 iterations. Each mini-batch contains both labeled and unlabeled data with a batch size of 4. A time-dependent coefficient is applied to contrastive loss: \( \lambda_c = 0.1 \times e^{4(1 - \frac{t}{t_{\text{max}}})^2} \), where \( t \) denotes the current iteration. For experiments on BraTS-2019, the schedule is extended to 20000 iterations, and the learning rate follows a polynomial decay: \( l_r = l_{r_{\text{base}}} \cdot \left(1 - \frac{t}{t_{\text{max}}}\right)^{0.9} \). To ensure consistency, all baseline methods are retrained under the same configuration. The experiments are conducted on a Windows 11 system equipped with an Intel 13900KF CPU, 128 GB RAM, and an NVIDIA L40 GPU.
We evaluated model performance using Dice Similarity Coefficient (DSC), Jaccard Index, 95th percentile Hausdorff Distance (HD95), and Average Surface Distance (ASD).
\subsection{Results} 
Quantitative comparisons between the proposed method and SOTA semi-supervised segmentation approaches are presented in Table~\ref{tab:LAcomparison} (LA), Table~\ref{tab:PANCREAScomparison} (Pancreas-CT), and Table~\ref{tab:BRAINcomparison} (BraTS-2019). Each of these Tables (Table~\ref{tab:LAcomparison},~\ref{tab:PANCREAScomparison}, ~\ref{tab:BRAINcomparison}) is divided into three parts: the first summarizes the baseline performance using fully labeled data (denoted as “V-Net”), the second reflects outcomes under a 10\%/90\% labeled-to-unlabeled ratio, and the third demonstrates performance with 20\% labeled data.

\begin{figure*}
    \centering
    \includegraphics[width=0.9\linewidth]{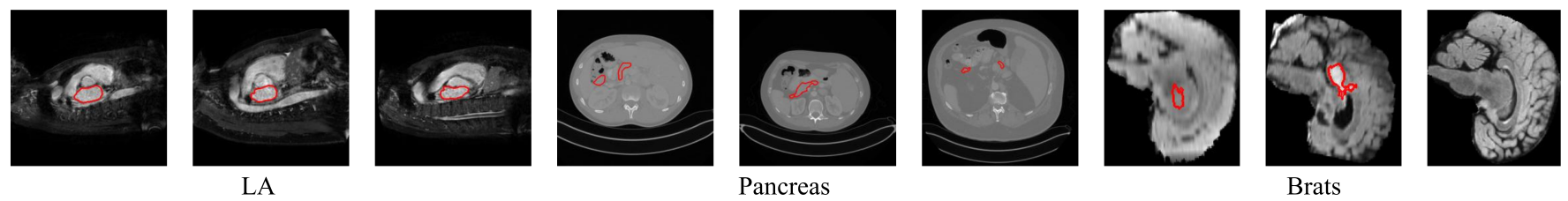}
    \caption{{Representative axial middle slices from LA, Pancreas-CT, and BraTS-2019 datasets. Compared with LA and BraTS-2019, the pancreas region exhibits larger anatomical variability and lower boundary contrast, complicating the segmentation task.}}
    \label{LA_PANCREAS_BRATS}
\end{figure*}

We highlight the following findings: 1) our method consistently achieves SOTA performance across all datasets under various partitioning settings. Specifically, on the LA dataset, with only eight labeled examples, our method yields a Dice Score of 89.95\%, surpassing the second best method, SS-Net \cite{wu2022exploring}, by 1.13\% in Dice Score and 1.78\% in Jaccard Score. 2) Even in the absence of labeled samples, our method matches the performance of the fully supervised V-Net. 3) Similar to the Pancreas-CT data set, the proposed method provides the highest segmentation accuracy, outperforming the second best method (MCF, UA-MT) by 1.71\% and 0.07\% in Dice Score when only 10\% and 20\% labeled data are available, respectively. 4) Considering the Brats-2019 dataset, the proposed model shows the highest Dice Score and Jaccard Score while also obtaining the lowest Hausdorff distance (95HD) under the labeled data settings 10\% and 20\%. In particular, in the 10\% labeled setting, our method surpasses AC-MT\cite{xu2023ambiguity} by a large margin of 1.07\% in Dice Score. Furthermore, under the 20\% labeled data setting, the proposed model achieves a Dice Score of 85.76\%, outperforming AC-MT of 84.63\%. While the proposed model does not outperform the compared methods in semi-supervised settings (see Table \ref{tab:LAcomparison}), it maintains competitive performance without increasing computational costs. {To explain this inconsistency, Figure~\ref{LA_PANCREAS_BRATS} visualizes representative axial middle slices from the three datasets. We observe that the pancreas exhibits substantial inter-patient variability in terms of shape, size, and anatomical location. Moreover, visceral fat surrounding the pancreas leads to contrast variation and often blurred or indistinct boundaries in CT scans, making accurate pseudo-labeling particularly challenging. In contrast, for the BraTS-2019 dataset, although brain tumors are highly heterogeneous, the annotated tumor regions are typically concentrated in only a few slices along the axial direction. This results in a more localized 3D label distribution, which simplifies segmentation and improves pseudo-label reliability under limited supervision.}

\begin{figure}
    \centering
    \includegraphics[width=0.9\linewidth]{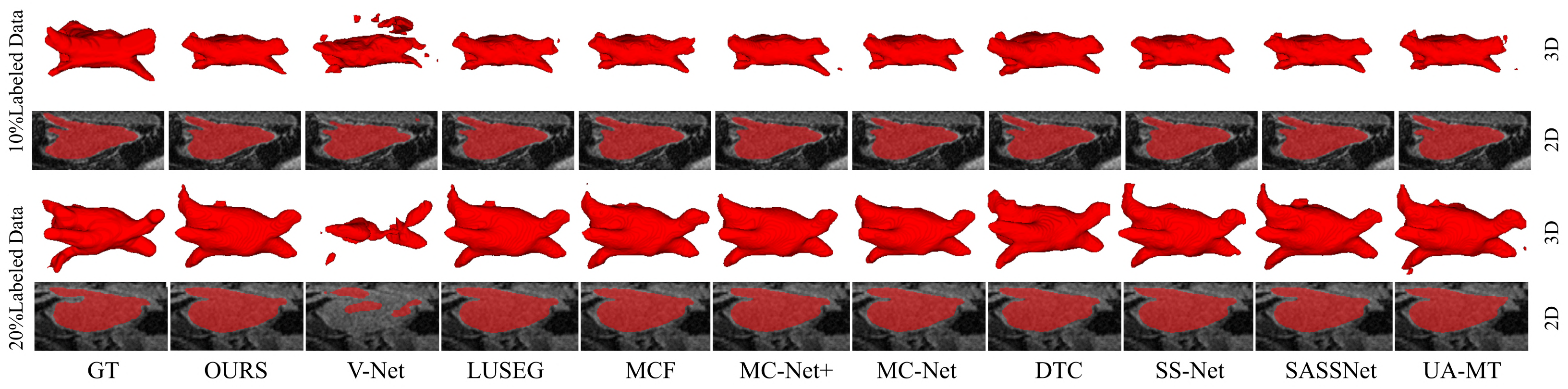}
    \caption{Visual comparison of segmentation results: from left to right, there are several exemplar results in 2D and 3D views obtained by different models on the LA dataset.}
    \label{fig:figLA}
\end{figure}

\begin{figure}
    \centering
    \includegraphics[width=0.9\linewidth]{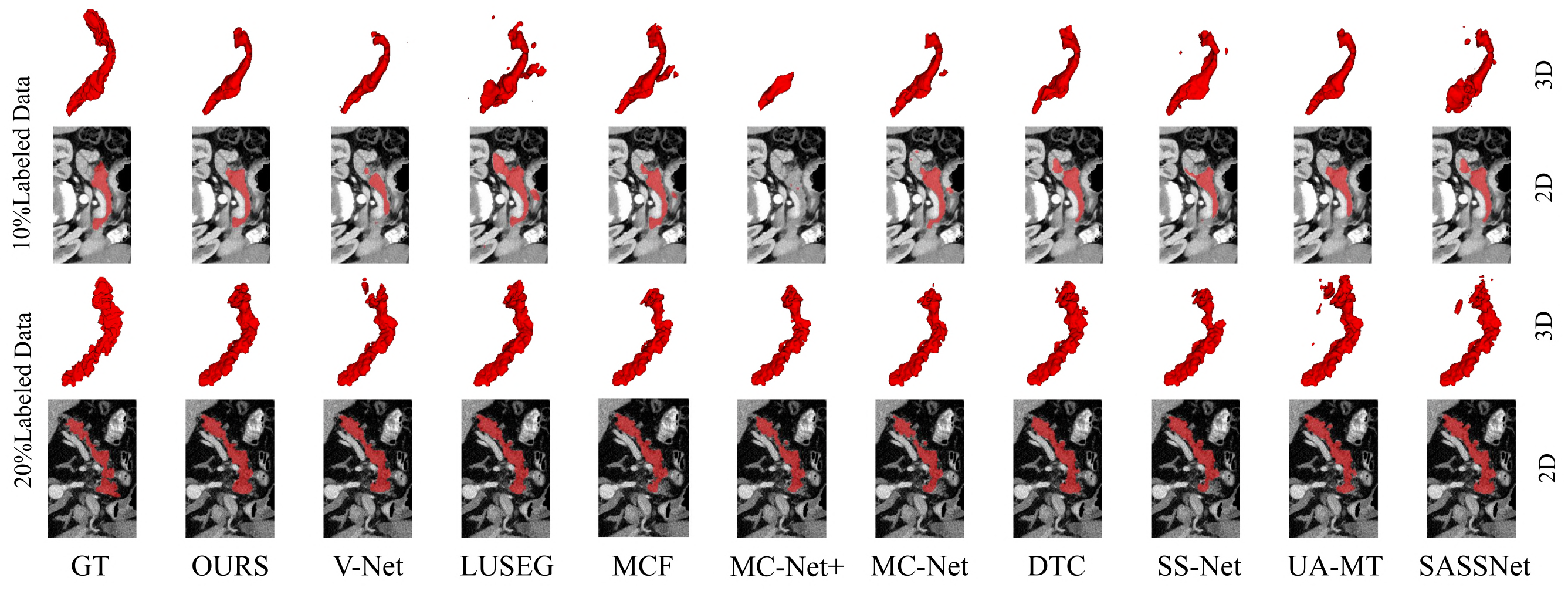}
    \caption{Segmentation results in 2D and 3D views are displayed from left to right, illustrating outputs from various models on the Pancreas dataset.}
    \label{fig:figPAN}
\end{figure}

\begin{figure}
    \centering
    \includegraphics[width=0.9\linewidth]{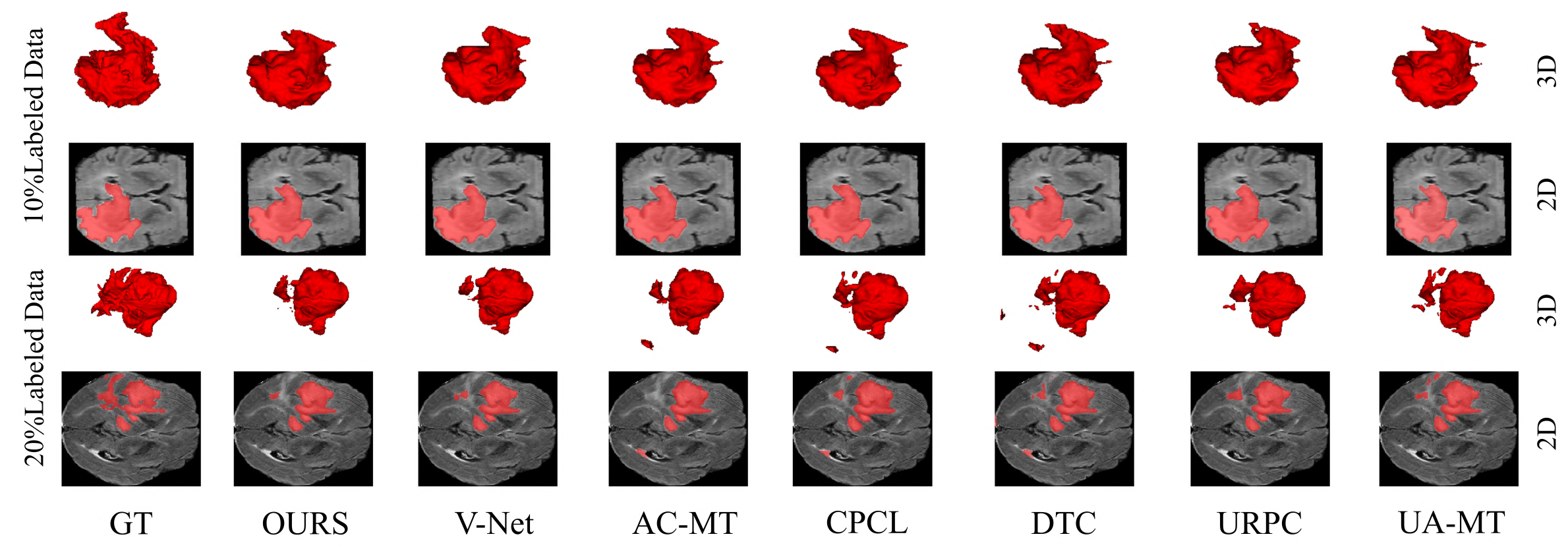}
    \caption{Segmentation results in 2D and 3D views obtained by different models on the brain tumor dataset.}
    \label{fig:figBrats}
\end{figure}

\begin{figure}[]
    \centering
    \includegraphics[width=0.9\linewidth]{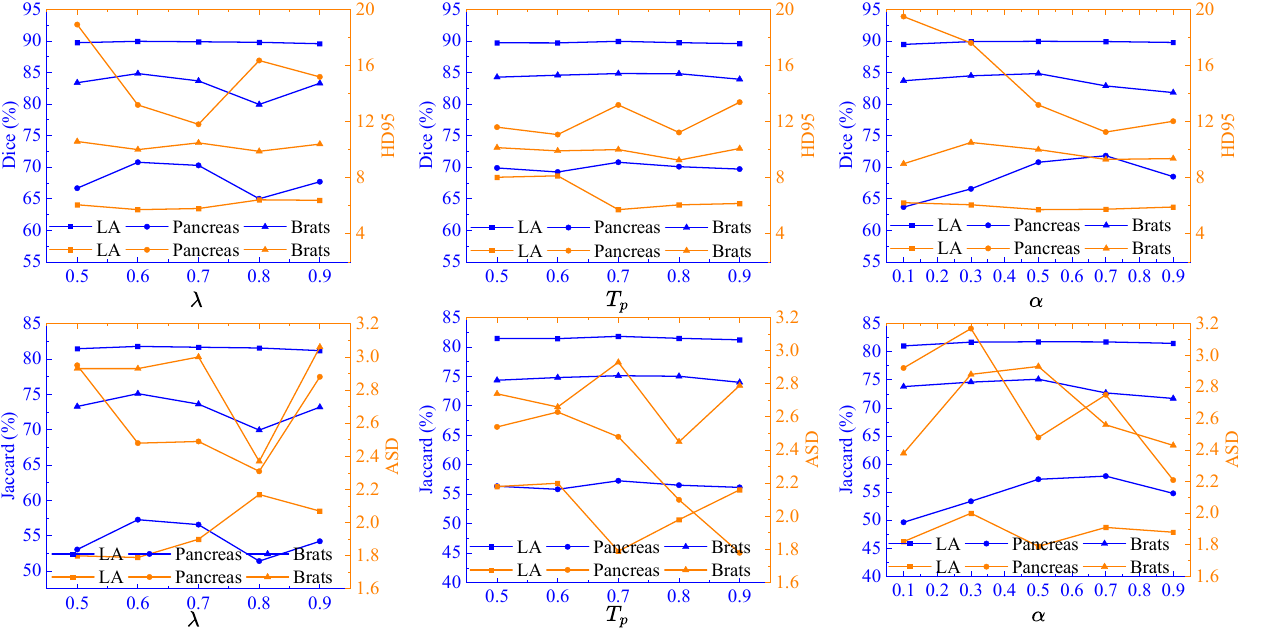}
        \label{fig:ablation-lambda}
    \caption{Performance metrics (Dice, Jaccard, ASD, HD95) for semi-supervised segmentation models in function with $\lambda$, $T_p$ and $\alpha$  using the three datasets (LA, Pancreas and Brats).}
    \label{fig:ablation-three}
\end{figure}

Figure~\ref{fig:figLA}, Figure~\ref{fig:figPAN}, and Figure~\ref{fig:figBrats} illustrate qualitative comparisons between existing segmentation approaches on the LA, Pancreas, and BraTS-2019 datasets, under 10\% and 20\% labeled training scenarios. Notably, our framework produces outputs that more closely resemble the ground truth, especially in terms of boundary precision and structural completeness. In addition, our approach provides a more comprehensive and detailed representation of the anatomical structures, particularly at the edges and branches, which are crucial features for clinical applications.

\vspace{-3pt}
\subsection{Ablation studies}

\subsubsection{Impact of each component} We validate the usefulness of each component on LA, Pancreas-CT and Brats-2019 datasets (trained with 10\% labeled data), as reported in Table \ref{Ablation}. We observed that: 1) the combination of all loss functions (CCE + PGL + UE + CR) achieves the best performance across all datasets, highlighting the mutual effects of these components. 2) Each individual component contributes to improving the segmentation performance of the model. 3) The PGL module has a substantial impact on the LA dataset but provides limited performance gain for the other two datasets. The UE module, on the other hand, shows minimal impact on the LA dataset. This can be attributed to the differences in dataset characteristics: Compared to LA, the Pancreas-CT dataset introduces higher segmentation complexity, is more challenging to segment, and is more likely to produce noisy pseudo-labels, while segmenting brain tumors is inherently more complex due to the uncertainty of tumor boundaries and the high heterogeneity of tumor appearance. In contrast, the LA dataset is relatively simple and may not require extensive uncertainty estimation. {Compared with using CPS or EFS alone, their combination (CCE) consistently leads to better results across all datasets, with CPS contributing more than EFS, particularly on the LA dataset.}

\subsubsection{Distance measurements}
{We replace the distance measurement (MSE) used in Eq. (\ref{EQ:6}) with KL divergence on three data sets, and the results are listed in Table \ref{tab:distance_measures}. We observed that while KL divergence remains an applicable solution for training models, MSE consistently provides better segmentation accuracy and boundary precision in the three datasets, especially in complex anatomical regions.}

\begin{table*}
\centering
\renewcommand{\arraystretch}{1.3}
\caption{Comparison of KL and MSE for $\mathcal{L}_{cr}$ (Eq.~(\ref{EQ:6})) under Different Label Ratios on LA, Pancreas-CT, and BraTS-2019.}
\label{tab:distance_measures}
\small
\resizebox{\textwidth}{!}{
\begin{tabular}{c|c|cccc|cccc|cccc}
\hline
\multirow{2}{*}{Labeled / Unlabeled} & \multirow{2}{*}{$\mathcal{L}_{cr}$} & \multicolumn{4}{c|}{LA} & \multicolumn{4}{c|}{Pancreas-CT} & \multicolumn{4}{c}{Brats-2019} \\
\cline{3-14}
&  & Dice (\%)  & Jaccard (\%)  & 95HD (voxel)  & ASD (voxel)  & Dice (\%)  & Jaccard (\%)  & 95HD (voxel)  & ASD (voxel)  & Dice (\%)  & Jaccard (\%) & 95HD (voxel) & ASD (voxel)  \\
\hline
\multirow{2}{*}{10\% / 90\%} 
& KL  & 89.01 ± 0.80 & 80.37 ± 1.27 & 6.82 ± 0.86 & 1.81 ± 0.20 & 26.33 ± 4.72 & 16.91 ± 3.26 & 50.65 ± 12.32 & 4.80 ± 2.43 & 78.82 ± 2.68 & 68.95 ± 2.99 & 10.81 ± 1.54 & \textbf{2.44 ± 0.39} \\
& MSE & \textbf{89.95 ± 0.58} & \textbf{81.81 ± 0.95} & \textbf{5.73 ± 0.63} & \textbf{1.79 ± 0.24} & \textbf{71.83 ± 4.15} & \textbf{57.87 ± 3.97} & \textbf{11.26 ± 3.57} & \textbf{2.75 ± 0.72} & \textbf{84.84 ± 1.45} & \textbf{75.13 ± 1.94} & \textbf{10.01 ± 1.70} & 2.93 ± 0.54 \\
\hline
\multirow{2}{*}{15\% / 85\%} 
& KL  & 86.56 ± 1.48 & 76.86 ± 2.15 & 8.95 ± 1.33 & 1.97 ± 0.26 & 53.18 ± 5.67 & 40.10 ± 5.09 & 34.05 ± 8.86 & 3.23 ± 1.08 & \textbf{84.83 ± 1.61} & \textbf{75.41 ± 2.09}& \textbf{8.47 ± 1.34} & \textbf{2.27 ± 0.37} \\
& MSE & \textbf{90.09 ± 0.61} & \textbf{82.08 ± 0.99} & \textbf{5.68 ± 0.53} & \textbf{1.81 ± 0.15} & \textbf{73.13 ± 3.75} & \textbf{59.85 ± 3.81} & \textbf{10.28 ± 2.16} & \textbf{1.86 ± 0.28} & 84.68 ± 1.63 & 75.23 ± 2.13 & 10.18 ± 1.85 & 2.85 ± 0.60 \\
\hline
\multirow{2}{*}{20\% / 80\%} 
& KL  & 90.63 ± 0.60 & 82.97 ± 0.98 & 5.52 ± 0.61 & 1.69 ± 0.18 & 66.14 ± 4.78 & 53.03 ± 4.67 & 19.61 ± 5.20 & 2.43 ± 0.57 & 84.59 ± 1.69 & 75.20 ± 2.18 & 8.96 ± 1.54 & 2.31 ± 0.41 \\
& MSE & \textbf{90.90 ± 0.54} & \textbf{83.41 ± 0.90} & \textbf{5.28 ± 0.58} & \textbf{1.63 ± 0.15} & \textbf{77.07 ± 3.86} & \textbf{64.35 ± 3.93} & \textbf{9.29 ± 1.79} & \textbf{1.73 ± 0.14} & \textbf{85.76 ± 1.41} & \textbf{76.51 ± 1.91} & \textbf{7.46 ± 1.09} & \textbf{1.99 ± 0.32} \\
\hline
\multirow{2}{*}{25\% / 75\%} 
& KL  & 91.20 ± 0.45 & 83.88 ± 0.75 & 5.42 ± 0.60 & 1.59 ± 0.15 & 63.90 ± 4.92 & 50.35 ± 4.77 & 19.57 ± 5.51 & 1.95 ± 0.27 & 83.68 ± 1.80 & 74.05 ± 2.28 & 7.47 ± 1.02 & 1.83 ± 0.30 \\
& MSE & \textbf{91.36 ± 0.44} & \textbf{84.15 ± 0.73} & \textbf{4.83 ± 0.40} & \textbf{1.56 ± 0.11} & \textbf{75.85 ± 2.67} & \textbf{62.44 ± 3.14} & \textbf{9.22 ± 1.67} & \textbf{1.60 ± 0.19} & \textbf{85.28 ± 1.52} & \textbf{75.94 ± 2.03} & \textbf{6.91 ± 0.95} & \textbf{1.75 ± 0.30} \\
\hline
\end{tabular}}
\end{table*}

\begin{table*}
\centering
\renewcommand{\arraystretch}{1.3}
\caption{{Comparison of the original contrastive loss $\mathcal{L}_c$ (Eq.~\ref{EQ:contras}) and the contrastive loss $\mathcal{L}_c$ without the third term ($\mathcal{L}_c\, \, \text{w/o}\, \, d(c_f, c_b)$) on three datasets under different labeled/unlabeled ratios. Best results are highlighted in bold.}}
\label{tab:lc}
\small
\resizebox{\textwidth}{!}{
\begin{tabular}{c|c|cccc|cccc|cccc}
\hline
\multirow{2}{*}{Labeled / Unlabeled} & \multirow{2}{*}{Loss type} & \multicolumn{4}{c|}{LA} & \multicolumn{4}{c|}{Pancreas-CT} & \multicolumn{4}{c}{Brats-2019} \\
\cline{3-14}
&  & Dice (\%)  & Jaccard (\%)  & 95HD (voxel)  & ASD (voxel)  & Dice (\%)  & Jaccard (\%)  & 95HD (voxel)  & ASD (voxel)  & Dice (\%)  & Jaccard (\%) & 95HD (voxel) & ASD (voxel)  \\
\hline
\multirow{2}{*}{10\% / 90\%} 
&$\mathcal{L}_c$  & \textbf{89.95 ± 0.58} & \textbf{81.81 ± 0.95} & \textbf{5.73 ± 0.63} & \textbf{1.79 ± 0.24} & \textbf{71.83 ± 4.15} & \textbf{57.87 ± 3.97} & \textbf{11.26 ± 3.57} & \textbf{2.75 ± 0.72} & \textbf{84.84 ± 1.45} & \textbf{75.13 ± 1.94} & \textbf{10.01 ± 1.70} & 2.93 ± 0.54 \\
& $\mathcal{L}_c\, \, \text{w/o}\, \, d(c_f, c_b)$  &89.82 ± 0.57&81.62 ± 0.93&5.98 ± 0.63&1.81 ± 0.15&62.91 ± 2.38 & 46.83 ± 2.72 & 34.04 ± 2.44 & 11.89 ± 0.91&83.95 ± 1.53 & 73.91 ± 2.01 & 10.33 ± 1.69 & 2.94 ± 0.48\\

\hline

\multirow{2}{*}{20\% / 80\%} 
&$\mathcal{L}_c$  & \textbf{90.90 ± 0.54} & \textbf{83.41 ± 0.90} & \textbf{5.28 ± 0.58} & \textbf{1.63 ± 0.15} & \textbf{77.07 ± 3.86} & \textbf{64.35 ± 3.93} & \textbf{9.29 ± 1.79} & \textbf{1.73 ± 0.14} & \textbf{85.76 ± 1.41} & \textbf{76.51 ± 1.91} & \textbf{7.46 ± 1.09} & \textbf{1.99 ± 0.32} \\
& $\mathcal{L}_c\, \, \text{w/o}\, \, d(c_f, c_b)$  &90.47 ± 0.72 & 82.75 ± 1.15 & 5.40 ± 0.64 & 1.73 ± 0.20&77.44 ± 2.05 & 64.06 ± 2.61 & 15.90 ± 3.26 & 4.73 ± 0.92&84.60 ± 1.62 & 75.09 ± 2.12 & 9.30 ± 1.62 & 2.51 ± 0.47\\
\hline

\end{tabular}}
\end{table*}

\begin{table*}[ht]
    \centering
    \renewcommand{\arraystretch}{1.3}
    \caption{{Ablation study on adaptive entropy thresholds $\tau_A$, $\tau_B$ computed via $\gamma^{\text{th}}$ percentiles.}}
    \label{tab:tau}
    \small
    \resizebox{\textwidth}{!}{
    \begin{tabular}{c|cccc|cccc|cccc}
        \hline
        \multirow{2}{*}{$\gamma^{\text{th}}$} & \multicolumn{4}{c|}{LA} & \multicolumn{4}{c|}{Pancreas-CT} & \multicolumn{4}{c}{Brats-2019} \\
        \cline{2-13}
        & \textbf{Dice (\%)}  & \textbf{Jaccard (\%)}  & \textbf{95HD (voxel)} & \textbf{ASD (voxel)} 
        & \textbf{Dice (\%)}  & \textbf{Jaccard (\%)} & \textbf{95HD (voxel)}  & \textbf{ASD (voxel)}  
        & \textbf{Dice (\%)}  & \textbf{Jaccard (\%)}  & \textbf{95HD (voxel)}  & \textbf{ASD (voxel)}  \\
        \hline
        10  & 89.57 ± 0.53 & 81.19 ± 0.87 & 6.11 ± 0.49 & 2.05 ± 0.18 & 70.53 ± 4.17 & 56.96 ± 3.93 & 11.92 ± 1.98 & 3.48 ± 0.96 & 81.39 ± 2.07 & 71.24 ± 2.52 & 9.82 ± 1.50 & 2.54 ± 0.40 \\
        20  & 89.79 ± 0.58 & 81.57 ± 0.95 & 5.92 ± 0.59 & 1.89 ± 0.17 & 68.29 ± 4.42 & 54.69 ± 4.27 & 12.59 ± 1.95 & 2.88 ± 0.53 & 81.39 ± 2.07 & 71.24 ± 2.52 & 9.82 ± 1.50 & 2.54 ± 0.40 \\
        30  & 89.70 ± 0.58 & 81.42 ± 0.94 & 5.93 ± 0.50 & 2.05 ± 0.24 & 67.79 ± 4.50 & 54.11 ± 4.23 & 16.69 ± 4.92 & 2.95 ± 0.77 & 84.84 ± 1.45 & 75.13 ± 1.94 & 10.01 ± 1.70 & 2.93 ± 0.54 \\
        40  & 89.83 ± 0.54 & 81.62 ± 0.89 & 6.07 ± 0.55 & 2.06 ± 0.23 & 70.00 ± 3.79 & 56.02 ± 3.78 & 13.86 ± 2.61 & 3.12 ± 0.62 & 84.31 ± 1.50 & 74.41 ± 2.00 & 10.14 ± 1.71 & 2.76 ± 0.47 \\
        50  & 89.72 ± 0.59 & 81.47 ± 0.97 & 6.11 ± 0.63 & 2.05 ± 0.22 & 65.69 ± 4.80 & 52.04 ± 4.43 & 15.39 ± 3.10 & 3.62 ± 1.34 & 83.83 ± 1.59 & 73.87 ± 2.09 & 9.96 ± 1.62 & 2.77 ± 0.46 \\
        60  & 89.41 ± 0.62 & 80.96 ± 1.00 & 6.45 ± 0.65 & 2.08 ± 0.18 & 69.00 ± 4.36 & 55.41 ± 4.18 & 12.10 ± 2.29 & 2.46 ± 0.30 & 84.50 ± 1.43 & 74.60 ± 1.94 & 8.32 ± 1.17 & 2.37 ± 0.38 \\
        70  & 89.95 ± 0.58 & 81.81 ± 0.95 & 5.73 ± 0.63 & 1.79 ± 0.24 & 65.76 ± 4.69 & 51.99 ± 4.35 & 15.93 ± 3.28 & 2.70 ± 0.63 & 83.87 ± 1.64 & 74.02 ± 2.13 & 9.21 ± 1.41 & 2.46 ± 0.41 \\
        80  & 89.69 ± 0.56 & 81.41 ± 0.93 & 5.91 ± 0.60 & 1.85 ± 0.14 & 71.83 ± 4.15 & 57.87 ± 3.97 & 11.26 ± 3.57 & 2.75 ± 0.72 & 84.52 ± 1.47 & 74.67 ± 1.96 & 10.34 ± 1.73 & 2.94 ± 0.49 \\
        90  & 89.89 ± 0.53 & 81.71 ± 0.87 & 6.07 ± 0.55 & 2.09 ± 0.21 & 70.06 ± 3.91 & 56.17 ± 3.80 & 12.27 ± 1.84 & 2.81 ± 0.75 & 84.01 ± 1.51 & 74.00 ± 2.03 & 9.34 ± 1.45 & 2.56 ± 0.43 \\
        \hline
    \end{tabular}}
\end{table*}

\subsubsection{Analysis based on hyper-parameters}  


We investigated the impact of hyperparameters $\lambda$, $\alpha$, and $T_p$ in the three datasets (LA, Pancreas-CT, Brats-2019). The parameter $\lambda$ controls the weight of $M$, as shown in Equation \ref{lambdaM}, influencing the strength of CR. In Figure \ref{fig:ablation-three}(a), for the LA dataset, the highest Dice and Jaccard scores and the lowest HD95 and ASD values are achieved when $\lambda$ is 0.6. For the Pancreas and Brats datasets, the highest Dice and Jaccard scores are also achieved at $\lambda$ = 0.6, but the HD95 and ASD values do not reach their lowest points. When $\lambda$ exceeds 0.6, both Dice and Jaccard scores show a noticeable decline across all three datasets, this indicates that for 3D datasets, a higher threshold might excessively filter out predictions, leading to information loss and diminished segmentation accuracy. Consequently, $\lambda$ is set to 0.6 by default. 

Moreover, we evaluate the usefulness of $T_p$, which regulates temperature scaling in the unsupervised loss term $L_p$ (\ref{losstp}). As illustrated in Figure \ref{fig:ablation-three}(b), the optimal performance is observed at $T_p$ = 0.5, both Dice and Jaccard metrics exhibit degradation when $T_p$ deviates from the optimal value of 0.5. The selected $T_p = 0.5$ optimally balances information retention and prevents class overlap. For $\alpha$, it controls the weight of the regularization term $L_{cr}$ (\ref{eq:supervised_loss}), affecting the stability of the model. As shown in Figure \ref{fig:ablation-three}(c), optimal performance is achieved with $\alpha$ = 0.5 for LA and Brats, and $\alpha$ = 0.6 for Pancreas. Simpler datasets like LA and Brats require less regularization, while more complex datasets like Pancreas benefit from stronger regularization to avoid overfitting and improve generalization. These settings are used in all our training configurations.

\subsubsection{Ablation on Contrastive Loss Variants} 
{Table~\ref{tab:lc} reports the test metrics using the original contrastive loss $\mathcal{L}_c$ and the contrastive loss $\mathcal{L}_c$ without the third term ($d(c_f,c_b)$). We observe that the original formulation $\mathcal{L}_c$ consistently achieves comparable or better performance across all datasets and labeled ratios. These results suggest that the third term effectively acts as a soft constraint, preventing the prototypes from diverging excessively and leading to a more robust and generalized representation.}

\subsubsection{Entropy thresholds in EFS masking}
{
To evaluate the robustness of our entropy-based filtering strategy (EFS), we perform experiments using various $\gamma^{\text{th}}$ values on LA, Pancreas-CT and Brats-2019 datasets. Table~\ref{tab:tau} reports the results on LA, Pancreas-CT, and BraTS-2019 datasets under different percentile settings \( \gamma \in \{10, 20, \dots, 90\} \), where the thresholds are computed as the \( \gamma\% \) percentile of per-batch entropy distributions. We observe that the model maintains relatively stable performance across a wide range of \( \gamma \) values, demonstrating the robustness of our uncertainty-aware filtering mechanism. We note that an extremely lower thresholds (e.g., \( \gamma \) =10\%) exclude too many informative voxels, while a higher threshold (e.g., \( \gamma \) =90\%) allows excessive noisy labels to participate in training. The optimal trade-off occurs at \( \gamma \) =70\% for LA, \( \gamma \) =80\% for Pancreas-CT, and \( \gamma \) =30\% for Brats-2019. These dataset-specific values are adopted as defaults throughout our experiments. These result suggest that the proposed EFS loss is not overly sensitive to the specific choice of \( \gamma \), and that percentile-based adaptive entropy masking provides a robust and effective way to stabilize training procedure.}

\noindent\textbf{Limitations.} While the model provides feasible performance in the 10\% labeled ratio setting, it still struggles to achieve optimal performance in terms of Average Symmetric Surface Distance (ASD) across datasets. To address this, we will incorporate loss functions that penalize errors near object boundaries and leveraging multi-scale features. Additionally, we will explore replacing the model framework with a Transformer architecture and conduct a comparative study between Transformer and CNN performance. Furthermore, we will consider a one-shot setting as suggested in \cite{pachetti2024systematic} for evaluation. Moreover, the current design prioritizes segmentation accuracy over parameter efficiency. Future research could explore lightweight variants of our architecture for edge deployments.

\vspace{-3pt}
\section{CONCLUSION}\vspace{-3pt}\label{CONCLUSION}

This study proposed a deep 3D input-based model for SSL based medical image segmentation. Specifically, we introduced an enhanced CL technique, equipped with consistency regularization and uncertain estimation to improve the quality of the pseudo-labels. Experimental results on three publicly medical datasets demonstrate that our approach yields reasonable performance with acceptable computation overhead. In future work, we will explore medical image segmentation in cross-domain situations and design a domain adaptation module \cite{10835760} to solve domain shifts.


\section*{Acknowledgments}
This research was funded by the National Natural Science Foundation of China grant number 82260360, and the Guangxi Science and Technology Base and Talent Project (2022AC18004, 2022AC21040). 

{
\bibliographystyle{unsrt}
\bibliography{cas-refs}
}

\end{document}